\documentclass{article}

\usepackage{microtype}
\usepackage{graphicx}
\usepackage{subfigure}
\usepackage{booktabs} 
\usepackage{amsmath}
\usepackage{amsthm,amssymb}
\usepackage{amsfonts}
\usepackage{bbm}
\usepackage{dcolumn}
\usepackage{hyperref}

\usepackage{cancel}

\usepackage{todonotes}
\usepackage{algorithm}
\usepackage{algpseudocode}
\usepackage[accepted]{icml2020}
\renewcommand{\S}{\mathcal{S}}
\newcommand{\A}{\mathcal{A}}

\newcommand{\Real}{\mathbb{R}}

\newcommand{\deq}{\mathrel{\mathop{:}}=}
\newcommand{\commentout}[1]{}

\newcommand{\norm}[1]{\left\lVert#1\right\rVert}
\DeclareMathOperator*{\argmin}{arg\,min}

\DeclareMathOperator{\diag}{diag}

\newtheorem{definition}{Definition}[section]
\newtheorem{theorem}{Theorem}[section]
\newtheorem{prop}{Proposition}[section]

\newtheorem{lemma}{Lemma}[section]
\newtheorem{fact}{Fact}[section]

\usepackage{listings}

\icmltitlerunning{TDprop}

\begin{document}

\twocolumn[
\icmltitle{TDprop: Does Jacobi Preconditioning Help Temporal Difference Learning?}



\icmlsetsymbol{equal}{*}

\begin{icmlauthorlist}
\icmlauthor{Joshua Romoff}{mila,mcgill}
\icmlauthor{Peter Henderson}{stanford}
\icmlauthor{David Kanaa}{mila,epm}
\icmlauthor{Emmanuel Bengio}{mila,mcgill}
\icmlauthor{Ahmed Touati}{mila,udem}
\icmlauthor{Pierre-Luc Bacon}{mila,udem}
\icmlauthor{Joelle Pineau}{mila,mcgill,fb}
\end{icmlauthorlist}

\icmlaffiliation{mila}{Mila}
\icmlaffiliation{mcgill}{McGill University}
\icmlaffiliation{stanford}{Stanford University}
\icmlaffiliation{fb}{Facebook AI Research}
\icmlaffiliation{epm}{\'Ecole Polytechnique de Montr\'eal}
\icmlaffiliation{udem}{Universit\'e de Montr\'eal}
\icmlcorrespondingauthor{Joshua Romoff}{joshua.romoff@mail.mcgill.ca}

\icmlkeywords{reinforcement learning, adaptive optimization, Jacobi preconditioning}

\vskip 0.3in
]



\printAffiliationsAndNotice{}  

\begin{abstract}
We investigate whether Jacobi preconditioning, accounting for the bootstrap term in temporal difference (TD) learning, can help boost performance of adaptive optimizers. Our method, TDprop, computes a per parameter learning rate based on the diagonal preconditioning of the TD update rule. 
We show how this can be used in both $n$-step returns and TD($\lambda$). Our theoretical findings demonstrate that including this additional preconditioning information is, surprisingly, comparable to normal semi-gradient TD if the optimal learning rate is found for both via a hyperparameter search.
In Deep RL experiments using Expected SARSA, TDprop meets or exceeds the performance of Adam in all tested games under near-optimal learning rates, but a well-tuned SGD can yield similar improvements -- matching our theory.
Our findings suggest that Jacobi preconditioning may improve upon typical adaptive optimization methods in Deep RL, but despite incorporating additional information from the TD bootstrap term, may not always be better than SGD.
\end{abstract}

\section{Introduction}
Reinforcement Learning (RL) systems are tasked with maximizing the cumulative sum of discounted rewards in a particular environment. In order to do so, most RL methods, 
rely on estimating the value function: the expected sum of discounted rewards. 
Estimating the value function efficiently, in terms of number of interactions with the environment, is crucial to the overall sample efficiency of the system. 
Temporal difference (TD)~\citep{sutton1988learning} attempts to improve efficiency of estimation by bootstrapping off of its own estimator. However, the use of this bootstrapping term requires optimizers which can handle non-\emph{iid} data, shifting distributions, and large stochasticity. These differences between supervised learning and TD learning, can have a major impact in terms of optimization. 

One approach to overcoming these challenges is to use an adaptive per-parameter learning rate, as existing adaptive optimizers do \citep{rmsprop, kingma2014adam}.
Most adaptive optimizers, however, are built with supervised learning in mind and do not explicitly account for the TD case.
Previous work has investigated whether adaptive optimizers can be constructed that are better suited for TD learning~\citep{henderson2018did,sun2020adaptive}.



We hypothesize that by taking into account the gradient of the bootstrap term in TD learning, we can build a more robust TD-specific adaptive optimizer. 
We follow recent advances \citep{devraj2017zap, chen2019zap} that derive the optimal gain (learning rate) matrix  from the stochastic approximation literature \citep{benveniste2012adaptive} for TD learning. Specifically, they find that in the linear case, the optimal gain matrix directly corresponds with the Least Squares TD method \citep{boyan1999least}. Instead, we propose to approximate the optimal gain matrix by its diagonal, the Jacobi preconditioner, which results in an efficient and principled adaptive method for TD -- building on prior work~\cite{givchi2015quasi, pan2017accelerated}. We theoretically compare the approach in the tabular setting against standard TD methods. We also show how this method can be easily adapted to the Deep RL setting and compare and contrast it with other deep learning optimizers. Surprisingly, we find that despite adding additional information from the bootstrap term, both theoretically and empirically, after a hyperparameter search TDprop behaves similarly to other optimizers (both TDProp and SGD meet or exceed the performance of Adam). 
This result suggests that while Jacobi preconditioning may be an improved approach to adaptive optimization in deep TD learning, further work is needed for adaptive optimization methods to yield a Pareto improvement over SGD.

\section{Preliminaries}

We aim to learn the value $v^\pi : \S \rightarrow \Real$ of a policy $\pi: \S \rightarrow \A$: 
\begin{equation}
    v^\pi(s) \deq \mathbb E ^\pi \left [ G_{t} \vert s_0 = s \right ] ,
\end{equation}
where $G_{t}\deq \sum_{k=0}^\infty \gamma^{k} r(s_{t+k}, a_{t+k})$ is discounted sum of future rewards starting at time-step $t$. 

We can train an estimate of the value function, $\hat v^\pi$, by regressing towards the $n$-step truncated $\lambda$-return, i.e, the truncated forward view:
\begin{equation}
\label{eq:truncated-td-lambda-return}
    G_{t:t+n}^{ \lambda} \deq \hat v^\pi (s_t) + \sum_{k=1}^n (\gamma \lambda)^{k-1} \delta_{t+k-1},
\end{equation}
where $\delta_{t} \deq  r(s_{t},a_{t}) +\gamma \hat v^\pi (s_{t+1}) - \hat v^\pi(s_t)$ is the one-step Temporal Difference (TD) error at time-step $t$ and $n $ is the truncation length. 

At each time-step $t$, the parameters of the value function, $\theta=(\theta_1, \theta_2, ..., \theta_m)$, can be updated as follows:
\begin{equation}
\label{eq:td_update}
    \theta_{t+1} = \theta_t + \alpha_{t+1} \delta_{t:t+n}^\lambda \nabla_\theta \hat v^\pi (x_t;\theta_t),
\end{equation}
where $\delta_{t:t+n}^\lambda=  G^\lambda_{t:t+n} - \hat v^\pi(s_t;\theta_t)$ is the error, $\alpha$ is the learning rate, and $x_t$ is the feature vector extracted at state $s_t$. 

\subsection{Stochastic Approximation}
The stochastic update procedure defined in Equation~\eqref{eq:td_update} can be seen as a specific case of stochastic approximation \citep{benveniste2012adaptive}:
\begin{equation}
    \theta_{t+1} = \theta_t + \alpha_{t+1} g(\theta_t, x_t), 
\end{equation}
where $\theta\in\Real^m$ is the parameter vector, $\alpha$ is the learning rate, and $g(\theta, x): (\theta \times x) \rightarrow \Real^m$ is the function that defines how the parameters are updated given observations $x$. 


The optimal gain matrix (learning rate), in terms of asymptotic convergence properties, is the negative of the inverse gradient of the expected update function \citep{benveniste2012adaptive}:
\begin{equation}
    H^{-1} = -(\nabla_\theta  \mathbb E^{\mu} \left [ g(\theta, x_t) \right ])^{-1},
\end{equation}
where $\mu$ is the sampling distribution and $H^{-1}\in\Real^{m\times m}$ is the resulting matrix gain.

The update to the parameters then becomes:
\begin{equation}
\label{eq:hessian_update}
    \theta_{t+1} = \theta_t + \alpha_{t+1} H^{-1}g(\theta_t, x_t), 
\end{equation}
where $H^{-1}$ is considered to be a preconditioner. Moreover, the choice of notation for $H$ is intentional, as in gradient descent $H$ corresponds with the \textit{Hessian} of the loss function. 


%

\subsection{Jacobi Preconditioning for Regression} 
\label{sec:approx-hessian}
In the following sections, we compare and contrast TD learning and supervised regression. To this end, we first present the common sum of squares error function:
\begin{equation}
    \mathcal{L}(\theta) = \mathbb E^{\mu} \left [ \frac{1}{2}( \hat y_t - y_t)^2 \right ],
\end{equation}
where $\hat y_t = f(x_t;\theta_t)$ is the estimate given the input $x_t$ and parameters $\theta_t$, and $y_t$ is the target at time $t$. 

The expected update direction is the negative gradient of $\mathcal L(\theta)$:
\begin{equation}
    \mathbb E^{\mu} \left [ g(\theta, x)\right ] = -\nabla_\theta \mathcal L(\theta) = -\mathbb E^{\mu} \left [  \delta_t \nabla_\theta \hat y_t \right ],
\end{equation}
where $\delta_t=\hat y_t - y_t$ is the error at time $t$.

The corresponding gradient of the update direction (i.e the Hessian of the loss function) is then:
\begin{align}
\label{eq:standard-hessian}
     H &= -\nabla_\theta( -\nabla_\theta \mathcal L(\theta)) = \nabla^2_\theta \mathcal L(\theta) \nonumber \\ 
     &= \mathbb E^{\mu} \left [ \nabla_\theta \hat y_t \nabla_\theta \hat y_t ^\top + \delta_t \nabla_\theta^2 \hat y_t \right ],
\end{align}
where $\nabla^2$ corresponds to applying the gradient operator twice.
Estimating the full Hessian can be computationally intractable due to the second order terms from Equation~\eqref{eq:standard-hessian}. Instead, the outer product approximation, also known as the Gauss Newton approximation,
drops the second order terms from Equation~\eqref{eq:standard-hessian}:
\begin{equation}
\label{eq:outer-product}
    H =  \nabla^2_\theta \mathcal L(\theta) \approx \mathbb E^{\mu} \left [ \nabla_\theta \hat y_t \nabla_\theta \hat y_t ^\top \right ].
\end{equation}
Finally, to obtain a per-parameter learning rate, we can approximate the Hessian matrix by its diagonal:
\begin{equation}
  \bar  H \approx \mathbb E^{\mu} \left [ \diag (\nabla_\theta \hat y_t \nabla_\theta \hat y_t ^\top) \right ],
  \label{eq:super_jacobi}
\end{equation}
which is known as the Jacobi preconditioner \citep{greenbaum1997iterative}. The main benefit of the diagonal approximation is that estimating and inverting the gain matrix (which is required to perform updates, see Equation~\eqref{eq:hessian_update}) is significantly cheaper computationally. The approximation accuracy will depend greatly on the problem at hand; nevertheless, both its low space and computational complexity has led to its usage \citep{lecun2012efficient}.  


\section{Jacobi Preconditioning for TD Learning}
\label{sec:tdprop_derivation}
We first recall that given the semi-gradient update function defined in Equation~\eqref{eq:td_update}, we have the following:
\begin{equation}
    g(\theta_t, x_t) =  \delta^\lambda_{t:t+n} \nabla_\theta \hat v^\pi (x_t;\theta_t),
\end{equation}
where $\delta^\lambda_{t:t+n} = G_{t:t+n}^\lambda - \hat v^\pi (s_t;\theta_t)$ is the error at time $t$. 

We set $\hat y_t = \hat{v}^\pi (s_t;\theta_t)$ and arrive at the following calculation for $H$:
\begin{align}
\label{eq:td_hessian}
     H &= - \nabla_\theta \mathbb E^{\mu} \left [   \delta^\lambda_{t:t+n}  \nabla_\theta \hat y_t  \right ] \nonumber \\
     & = - \mathbb E^{\mu} \left [ \nabla_\theta \delta^\lambda_{t:t+n}\nabla_\theta \hat y_t^\top  + \delta^\lambda_{t:t+n} \nabla^2_\theta \hat y_t    \right ].
\end{align}
To obtain an efficient adaptive optimizer we propose to use the diagonal approximation (the Jacobi preconditioner) as described in Equation~\eqref{eq:super_jacobi}:
\begin{equation}
\label{eq:outer_product_td}
   \bar H \approx - \mathbb E^{\mu} \left [ \diag(\nabla_\theta \delta^\lambda_{t:t+n}\nabla_\theta \hat y_t^\top ) \right ].
\end{equation}
To compare this expression to what was obtained for supervised regression in Equation~\eqref{eq:super_jacobi}, we can expand the outer product:
\begin{align}
     \bar H 
     &= \mathbb E^{\mu}  \bigg [ \diag \bigg ( \nabla_\theta \hat  y_{t}\nabla_\theta \hat y_t^\top -  \lambda^{n-1}\gamma^n \nabla_\theta \hat y_{t+n} \nabla_\theta \hat y_t^\top + \nonumber \\
     &\sum_{k=1}^{n-1} (\gamma \lambda)^{k-1} \left ( \gamma \lambda - \gamma \right ) \nabla_\theta \hat y_{t+k}\nabla_\theta \hat y_t^\top \bigg )\bigg],
\end{align}
where we note that the left most term $\nabla_\theta \hat y_{t}\nabla_\theta \hat y_t^\top$ is the same as the diagonal outer product approximation that arises from the sum of squares loss function in Equation~\eqref{eq:super_jacobi}. The remaining terms are unique to temporal difference learning. Moreover, the terms inside the summation disappear when $\lambda=1$ (i.e, when not using $\lambda$-returns).

\subsection{Theoretical Analysis}
We prove certain convergence properties of applying the Jacobi preconditioner to TD($0$), following a similar asymptotic analysis to \cite{schoknecht2003td} with constant step-sizes. The extension to TD($\lambda$) and $n$-step returns is provided in Appendix~\ref{app:yaymoretheory} with analogous results. We begin by noting that we aim to solve the following linear equation - assuming a tabular representation and uniform updates:
\begin{equation}
    r + (\gamma P - I) v= 0,
\end{equation}
where $r\in \mathbb R^{\lvert S \rvert}$ is the expected reward vector, $P \in  \mathbb R^{\lvert S \rvert \times \lvert S \rvert}$ is the transition matrix and $v \in \mathbb R^{\lvert S \rvert}$ is the estimated value function. We note that $r + (\gamma P - I) v= 0$ at the solution, i.e, when $v=v^*$. 

We solve for $v$ via the following iterative update:
\begin{align}
    v_{t+1} &= v_t - \alpha \left ( H v_t -  r \right ),
\end{align}
where $\alpha$ is the constant learning rate, $H =  \left (I - \gamma P \right )$, and $v_t$ is the estimated value function at time $t$.

By defining the error vector as $e_t = v_t - v^*$, where for $v^*$ we have that $Hv^* - r=0$, we can derive the following recursion:
\begin{align}
    e_{t+1} &= v_{t+1} - v^* \nonumber \\
    &= (I - \alpha H ) v_t + \alpha  r - v^* + \underbrace{\alpha(Hv^* - r)}_{\text{=0}} \nonumber \\
    &=  (I - \alpha H )(v_t - v^*) = (I - \alpha H )^{t+1}e_0.
\end{align}
Thus, the error at time-step $t$ depends on the initial error at time-step $0$ and the matrix $(I - \alpha H)$. 

One useful metric for measuring convergence speed is the asymptotic convergence rate, which we now define.
\begin{definition}{(asymptotic convergence rate)} Given the recursion of error vectors $ e_{t+1}= (I - \alpha H )^{t+1}e_0$, the asymptotic convergence rate is defined as:
$$
\lim_{t\rightarrow \infty} \max_{e_0 \in \mathbb R^{\lvert S \rvert} \backslash 0} \left(\frac{\norm{e_t}}{\norm{e_0}}\right) ^{\frac{1}{t}}= \rho(I - \alpha H),$$
\end{definition}
where $\rho(\cdot)$ is the spectral radius. 

Applying the Jacobi preconditioner to the original system we get the following iterative formula:
\begin{equation}
    v_{t+1} = v_t - \alpha \bar H ^{-1} \left ( H v_t -  r \right )
\end{equation}
where following Equation~\eqref{eq:outer_product_td}, we have $\bar H = \diag(H) = \diag(I-\gamma P)$. We also note that the asymptotic convergence rate of the preconditioned system is $\rho(I - \alpha \bar H^{-1} H) $.

Using the theory of regular splittings \cite{varga1969matrix} we can frame both the Jacobi preconditioner and the original system as regular splittings and thereby prove that it has a better convergence rate.
\begin{definition}{(regular splitting Definition 3.28 \cite{varga1969matrix})} \label{def:regular_icml}
    If $H=B-C$, $B^{-1} \geq 0$, and $C \geq 0$ for all components, then $B-C$ is said to be a regular splitting of $H$.
\end{definition}
Moreover, we have the following proposition that allows us to compare the asymptotic convergence rates of different regular splittings.
\begin{prop}{(comparing regular splittings Theorem 3.32 \cite{varga1969matrix}):}
\label{prop:spectral_comp_icml}
Let $(B_1, C_1)$, and $(B_2, C_2)$ be regular splittings of $H$. Then if $H^{-1}\geq 0$ and $0 \leq C_2 \leq C_1$ for all components, then:
\begin{equation}
    0 \leq \rho (B_2^{-1} C_2) \leq \rho( B_1^{-1} C_1) < 1.
\end{equation}
\end{prop}
Following Definition~\ref{def:regular_icml}, and using $H=I-\gamma P$, the Jacobi preconditioner can be seen as a regular splitting $H=\bar B - \bar C$ where $\bar B = \bar H$ and $\bar C = \bar B - H$. Similarly, for standard TD we have that $H=B-C$ where $B = I$ and $C = \gamma P$ forms a valid regular splitting of $H$. With both methods framed in terms of regular splittings, we can now compare their convergence rates using Proposition~\ref{prop:spectral_comp_icml} and setting the learning rate to $1$, i.e, the standard value iteration setting.
\begin{theorem}\label{the:convergence_icml} Let $H=I-\gamma P$ and $\bar H= \diag(H)$, then we have that:
\begin{equation}
    \rho(I - \bar H^{-1} H) \leq \rho(I - H) < 1.
\end{equation}
The proof is provided in Appendix~\ref{app:yaymoretheory}.
\end{theorem}
The previous theorem omitted the use of learning rates, in fact, it explicitly assumed a learning rate of $1$. However, it is common to perform a hyperparameter search over the learning rate to obtain the best possible performance. To this end, the optimal learning rate $\alpha^*$, for the case where $eig(H)\in\Real$, is derived in the following proposition.
\begin{prop} For a matrix $H$ with only positive real eigenvalues $eig(H)=\{\lambda_1, \lambda_2, ... \} \in \Real^{>0}$ we have that:
\begin{equation}
    \min_\alpha \rho(I-\alpha H) = \frac{\lambda_{\max}^H - \lambda_{\min}^H}{\lambda_{\max}^H + \lambda_{\min}^H}=\frac{\kappa (H) - 1}{\kappa(H) + 1}.
\end{equation}
\label{optimal_alpha_icml}
where $\kappa(H)=\frac{\lambda_{\max}^H}{\lambda_{\min}^H}$ is the condition number of $H$. 

The proof is provided in Appendix~\ref{app:yaymoretheory}.
\end{prop}


We highlight that from Proposition~\ref{optimal_alpha_icml}, the \emph{optimal spectral radius} is a monotonically increasing function of the \emph{condition number}, which means that poorly conditioned matrices will induce a slow convergence. As a result, we seek to reduce the condition number of $H$ with the Jacobi preconditioner. By comparing the condition numbers of the Jacobi preconditioned TD and standard TD we can determine which method has better convergence properties and performs best under their respective optimal learning rates in the case where $H$ is symmetric. 
\begin{theorem}
\label{the:jacobi_vs_td_icml}
Let $H=\left(I - \gamma P \right )$ and $\bar{H}=\diag\left(I - \gamma P \right)$, then assuming that $H$ is symmetric we have that:
\begin{equation}
    \kappa \left (\bar H^{-1} H \right ) \leq 2 \kappa \left (H \right ).
\end{equation}
The proof is provided in Appendix~\ref{app:yaymoretheory}.
\end{theorem}
In words, when $H$ is symmetric, the condition number of the Jacobi preconditioned system is at most a constant factor of $2$ worse than the original system. In practice, we would expect that in the worst case, once a hyperparameter search has been conducted over the learning rate, the Jacobi preconditioner would have similar performance to the original system.

\subsection{Practical Implementation}
We seek to track all required statistics for the diagonal outer product approximation, specifically (for each parameter $i$):
\begin{equation}
    \bar H ^{i,i} = z^i = - \mathbb{E}^{\mu_{\theta}} \left [ \nabla_{\theta_i} \delta^\lambda_{t:t+n}\nabla_{\theta_i} \hat y_t \right ].
\end{equation}

In practice, we use $\lvert \bar H \rvert$ because in non-convex optimization $H$ might be indefinite, see \citep{dauphin2015equilibrated}. Moreover, we found in initial testing that tracking $\bar H$ and then computing $\lvert \bar H \rvert$, led to poor performance due to the cancellation of positive and negative samples. Instead, to track $z$ we compute an exponential moving average of the squared sampled statistic:
\begin{equation}
    z_{t+1} = \beta z_t + (1-\beta) (- \nabla_{\theta} \delta^\lambda_{t:t+n} \odot \nabla_{\theta} \hat y_t )^2,
\end{equation}
where $\odot$ is the element-wise product and $\beta \in [0, 1)$ is the tracking hyperparameter.

We then update the parameter vector $\theta$ using the square root of $z$:
\begin{equation}
    \theta_{t+1} = \theta_t + \alpha_{t+1} \left ( Z_{t+1}^\frac{1}{2}  + \epsilon I \right )^{-1} \delta_{t:t+n}^{\lambda} \nabla_{\theta} \hat y_t , 
\end{equation}
where $Z_{t+1}$ is the diagonal matrix formed from the elements of the vector $z_{t+1}$, $\alpha$ is the global learning rate, and $\epsilon$ is a damping hyperparameter. 


\section{Experiments and Discussion}
\label{sec:experiments}

\begin{figure}
    \centering
        \includegraphics[width=0.49\textwidth]{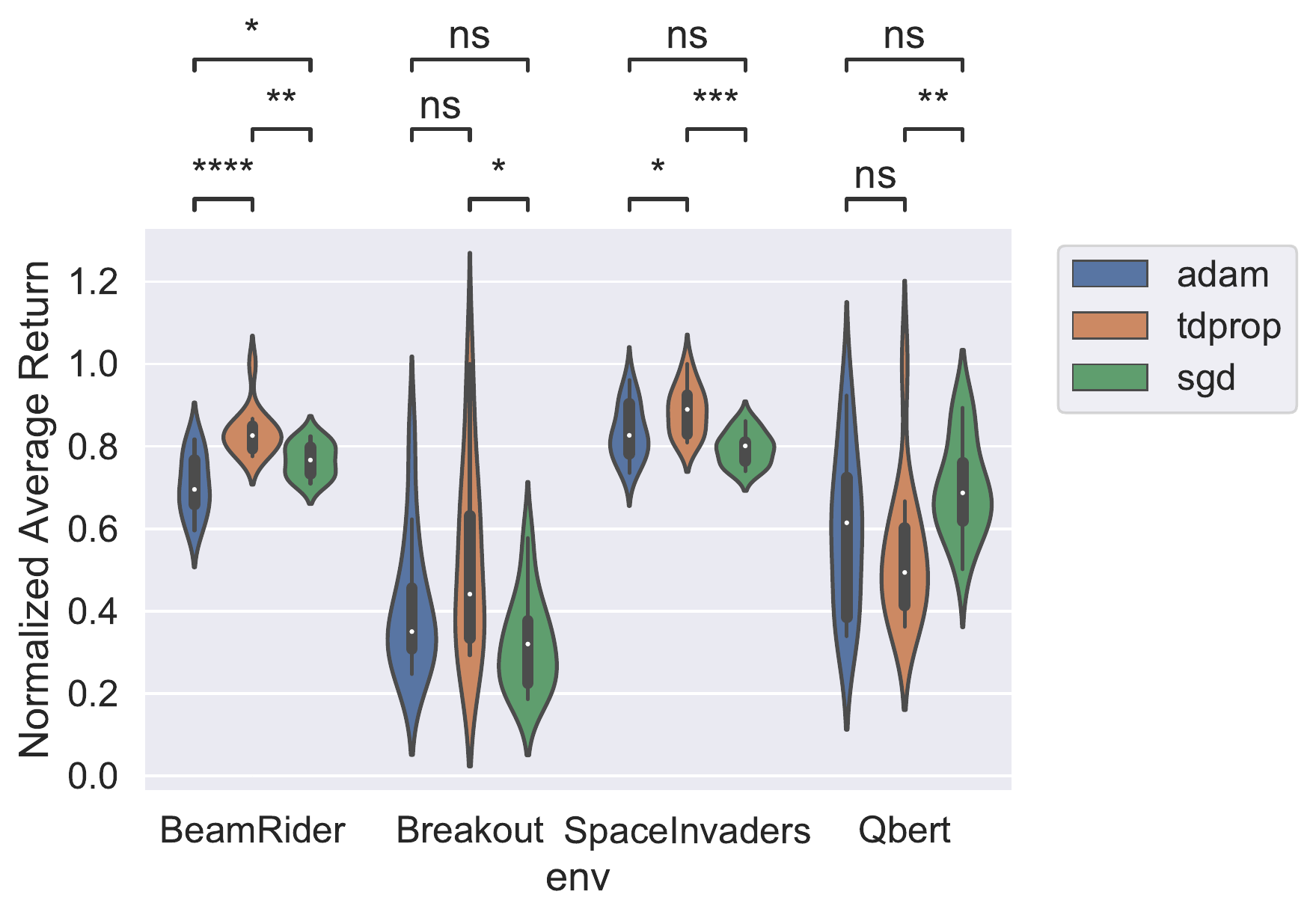}
    \caption{Top 25th percentile of hyperparameters by normalized average return. Significance testing using Welch's t-test. P-value annotation legend. ns: $0.05 < p <= 1$; *: $0.01 < p <= 0.05$; **: $0.001 < p <= 0.01$; ***: $0.0001 < p <=  0.001$; ****: $p <= 0.0001$. See Appendix~\ref{app:analysis} for more details and results.}
    \label{fig:violinplotmain}
\end{figure}

    
We compare TDprop to Adam~\citep{kingma2014adam}, as well as vanilla stochastic gradient descent (SGD), on four Deep RL tasks selected from the Arcade Learning Environment (ALE) \citep{bellemare2013arcade}: Beam Rider, Breakout, Qbert, and Space Invaders. We train each algorithm for $10$M training steps using $n$-step expected SARSA \citep{van2009theoretical}, modifying the A2C implementation of~\citet{pytorchrl}. Full experimental details and code references are provided in Appendix~\ref{app:experiment}.


Figure~\ref{fig:violinplotmain} shows the results of randomly sampling $50$ hyperparameter configurations. For a more in-depth discussion and extended results, see Appendix~\ref{app:analysis}. For the top 25th percentile of hyperparameters TDprop performs as well as or significantly better than Adam in all four games. 
However, confirming the theory of Theorem~\ref{the:jacobi_vs_td_icml}, we find that vanilla SGD under optimal learning rates performs as well as or better than Adam in all games tested and beats TDprop in one game -- in all cases coming close to the TDprop achieved performance.
Our results suggest that while TDProp improves performance by a small, but statistically significant, amount under a hyperparameter search in some settings, SGD can as well in other settings. The theory we derive provides some explanation for this phenomenon that we hope may lead to a better understanding of optimization in TD learning and future TD-specific optimizers.





\newpage
\bibliography{bib}
\bibliographystyle{icml2020}

\newpage
\appendix
\onecolumn

%

\section{Supplemental Materials}

\subsection{Carbon Impact Statement}

This work contributed a rough estimate of 3.772 kg of $\text{CO}_{2eq}$ to the atmosphere and used 125.718 kWh of electricity, having a CAN-specific social cost of carbon of \$-0.03 (\$-0.04, \$-0.02). This assumes 1153.377 hours of runtime, a region-specific carbon intensity of 30 g per kWh (see StatCan CANSIM Table 127-0002 for 2011-2015), a NVIDIA V100 PCIe GPU, a Core i7-5930K CPU, a CPU utilization of 0.6, a GPU utilization of 0.1. These values were based on A2C utilization of CPU performance as published in the Appendix logs of \citep{henderson2020towards} which we expect to be similar to our Expected SARSA implementation. The social cost of carbon uses models from \citep{ricke2018country} and this statement and carbon emissions information was generated using the \emph{get-rough-emissions-estimate} script of the \emph{experiment-impact-tracker}~\citep{henderson2020towards}. Note the Canada-specific social cost of carbon is negative in this case as explained by \citet{ricke2018country}: ``The CSCC captures the amount of marginal damage (or, if negative, the benefit) expected to occur in an individual country as a consequence of additional CO2 emission... Northern Europe, Canada and the Former Soviet Union have negative CSCC values because their current temperatures are below the economic optimum.''

\subsection{Extended Related Work}

\citet{devraj2017zap} derived and studied using the optimal gain matrix from stochastic approximation \citep{benveniste2012adaptive} for linear TD learning. They found that in the linear case, the optimal gain matrix directly corresponds with the Least Squares Temporal Difference (LSTD) method \citep{boyan1999least}. Recently, the approach was extended to the non-linear function approximation setting \citep{chen2019zap}. Unlike those methods, we propose to use a diagonal approximation of the gain matrix. This change provides us with a computationally tractable approach that can scale to millions of parameters, as is common in the Deep RL setting.

A wide range of work has examined adaptive optimization and preconditioners in supervised learning. For example, \citet{lecun2012efficient} describe the benefits of the diagonal preconditioner as well as efficient implementations. \citet{schaul2013no} propose a method for a adaptively tuning both the global learning rate as well as the per parameter learning rates based off of both the Jacobi preconditoner and local variance of the gradient. \citet{dauphin2015equilibrated} discuss trade-offs between the Jacobi preconditioner and the equilibriated preconditioner (which has similar properties to popular methods such as RMSprop~\citep{rmsprop} and Adam~\citep{kingma2014adam}). Finally, \citet{martens2014new} present a unified view of diagonal methods such as RMSprop and Adam for approximating the empirical Fisher matrix. Recently, \citet{sun2020adaptive} extended the adaptive update rule from \citet{duchi2011adaptive} to the TD setting and studied its convergence properties. Their theoretical results validate the use of standard adaptive optimizers from the Deep Learning literature in the TD setting. We compare and contrast our method to state of the art Deep Learning optimizers in Section~\ref{sec:experiments}.

There has been a vast array of work that explored adaptive optimizers and preconditioners for linear TD learning. For example, Scalar Incremental Delta-Bar-Delta (SID) \citep{dabney2014adaptive} extend Incremental Delta Bar Delta (IDBD) \citep{sutton1992adapting} to linear TD and adaptively tune a single global learning rate. Similarly, \citep{dabney2012adaptive} derive and examine an optimal global (not per parameter) learning rate for linear TD. Recently, TD Incremental Delta Bar Delta (TIDBD) \citep{kearney2019learning}, adaptively learn a per parameter learning rate based on the correlation between state features and TD errors. To our knowledge, however, TIDBD has not been extended to the non-linear setting with TD learning. In terms of preconditioners, \citet{yao2008preconditioned} present a generalized framework for using varying preconditioners in TD learning and subsequently in \citet{yaotemporal} propose to use the full matrix $H^{-1}$ as a preconditioner for linear TD learning. Perhaps the closest works to our own are approaches based on approximating $H$, as in \citet{givchi2015quasi, pan2017accelerated}.  Both works propose and examine the use of the diagonal approximation, however, in both cases the design of the algorithm and the empirical analysis is restricted to the linear setting. We expand on the theoretical linear analysis proposed in these works and provide empirical evidence in deep RL settings. In particular, our theoretical comparison of performance characteristics under optimal learning rates and matching experimental investigation aims to provide more ties between theory and practice under hyperparameter searches.


Finally, in the tabular case, Jacobi preconditioning can be interpreted as a per state learning rate based on a partial model of the world dynamics. Similarly, performing expected TD updates using a learned model of the transition dynamics has been shown to improve sample efficiency in both the tabular~\citep{sutton1991dyna} and the Deep RL setting \citep{feinberg2018model}. Unlike those methods, our approach does not plan with the learned model and only requires the tracking of a partial model of the dynamics, the probability of remaining in the same state.
\subsection{Extended Theoretical Analysis}
\label{app:yaymoretheory}
In this section, we prove certain convergence properties of applying the Jacobi preconditioner to TD learning, following a similar analysis to that of \citet{schoknecht2003td}. We begin by noting that we aim to solve the following linear equation, which assuming a tabular representation and uniform updates has the following form:
\begin{equation}
    r + (\gamma P - I) v= 0,
\end{equation}
where $r\in \mathbb R^{\lvert S \rvert}$ is the expected reward vector, $P \in  \mathbb R^{\lvert S \rvert \times \lvert S \rvert}$ is the transition matrix and $v \in \mathbb R^{\lvert S \rvert}$ is the estimated value function. We note that $r + (\gamma P - I) v= 0$ at the solution, i.e, when $v=v^*$. 

We seek to solve for $v$ via the following iterative update:
\begin{align}
    v_{t+1} &= v_t - \alpha \left ( H v_t -  r \right ),
\end{align}
where $\alpha$ is the learning rate, $H =  \left (I - \gamma P \right )$, and $v_t$ is the estimated value function at time $t$.

By defining the error vector as $e_t = v_t - v^*$, where for $v^*$ we have that $Hv^* - r=0$, we can derive the following recursion:
\begin{align}
    &e_{t+1} = v_{t+1} - v^* \nonumber \\
    &= (I - \alpha H ) v_t + \alpha  r - v^* + \underbrace{\alpha(Hv^* - r)}_{\text{=0}} \nonumber \\
    &=  (I - \alpha H )(v_t - v^*) = (I - \alpha H )^{t+1}e_0.
\end{align}
Thus, the error at time-step $t$ depends on the initial error at time-step $0$ and the matrix $(I - \alpha H)$. Specifically, to have convergence we require that the error goes to $0$ as $t\rightarrow \infty$. Which brings us to the definition of convergent matrices.
\begin{definition}{(convergent matrix: Definition 1.9 \citep{varga1969matrix})} Let $A \in \mathbb C ^{n \times n}$. Then $A$ is convergent (to zero) if the sequence of matrices $A, A^2, A^3, ...$ converges to the null matrix $0$, and is divergent otherwise.
\end{definition}

Conveniently, we have the following theorem that gives us the necessary and sufficient conditions for a matrix to be convergent.
\begin{theorem}{(convergence requirement: Theorem 1.10 \citep{varga1969matrix})} If $A\in \mathbb C ^{n \times n}$, then $A$ is convergent if and only if $\rho(A) < 1$, 
where $\rho(\cdot)$ is the spectral radius.
\end{theorem}
Thus, to have a convergent matrix we need certain conditions on the spectral radius, which we now formally define. 
\begin{definition}
\label{spectral_def}
The spectral radius of a matrix $A$ is defined as:
\begin{equation}
    \rho(A) = \max \left \{ \left \lvert \lambda^A_1 \right \rvert, \left \lvert \lambda^A_2 \right \rvert, ..., \left \lvert \lambda^A_m \right \rvert \right  \},
\end{equation}
where $\lambda^A$ are the eigenvalues of the matrix $A$.
\end{definition}
Moreover, we have the following theorem that tells us about the spectral radius of positive matrices, which will be useful later in the section.
\begin{theorem}{(perron: Theorem 10.2.4 from \cite{greenbaum1997iterative})} If $A \in \mathbb R ^{n \times n}$ and $A\geq0$ for all components, then $\rho(A)$ is an eigenvalue of $A$ and there is a nonnegative vector $v \geq 0$ with $\norm{v} = 1$, such that $Av=\rho(A)v$. 
\label{the:perron}
\end{theorem}
Thus, for positive matrices we have that the spectral radius corresponds to a simple eigenvalue. Hereafter, we will use $eig(A)$ as a shorthand to denote the eigenvalues of the matrix $A$.

The spectral radius is not only used to determine convergence but also to measure the asymptotic convergence speed.
\begin{definition}{(asymptotic convergence rate)} Given the recursion of error vectors $ e_{t}= (I - \alpha H )^{t}e_0$ and an arbitrary vector norm $\norm{v}$ with its compatible matrix norm $\norm{A}\deq \max_{v \in \mathbb R^{\lvert S \rvert} \backslash 0} \frac{Av}{\norm{v}}$, the asymptotic convergence rate is defined as:
\begin{align}
    \lim_{t\rightarrow \infty} \max_{e_0 \in \mathbb R^{\lvert S \rvert} \backslash 0} \left(\frac{\norm{e_t}}{\norm{e_0}}\right) ^{\frac{1}{t}} &= \lim_{t\rightarrow \infty} \max_{e_0 \in \mathbb R^{\lvert S \rvert} \backslash 0} \left(\frac{\norm{\left (I - \alpha H \right )^t e_0}}{\norm{e_0}}\right) ^{\frac{1}{t}}\nonumber \\
    &= \lim_{t\rightarrow \infty}  \left( \norm { \left (I - \alpha H \right ) ^t} \right) ^{\frac{1}{t}}\nonumber \\
    &= \rho(I - \alpha H)
\end{align}
\end{definition}

Applying the Jacobi preconditioner to the original system we get the following iterative formula:
\begin{equation}
    v_{t+1} = v_t - \alpha \bar H ^{-1} \left ( H v_t -  r \right )
\end{equation}
where following equation~\ref{eq:outer_product_td}, we have $\bar H = \diag(H) = \diag(I-\gamma P)$. We also note that the asymptotic convergence rate of the preconditioned system is $\rho(I - \alpha \bar H^{-1} H) $.

Using the theory of regular splittings \cite{varga1969matrix} we can frame both the Jacobi preconditioner and the original system as regular splittings and thereby prove that it has a better convergence rate.
\begin{definition}{(regular splitting: Definition 3.28 \cite{varga1969matrix})} \label{def:regular}
    If $H=B-C$, $B^{-1} \geq 0$, and $C \geq 0$ for all components, then $B-C$ is said to be a regular splitting of $H$.
\end{definition}
Moreover, we have the following proposition that allows us to compare the asymptotic convergence rates of different regular splittings.

\begin{prop}{(comparing regular splittings: Theorem 3.32 \cite{varga1969matrix} and Corollary 10.3.1 from \cite{greenbaum1997iterative}):}
\label{prop:spectral_comp}
Let $(B_1, C_1)$, and $(B_2, C_2)$ be regular splittings of $H$. Then if $H^{-1}\geq 0$ and $0 \leq C_2 \leq C_1$ for all components, then:
\begin{equation}
    0 \leq \rho (B_2^{-1} C_2) \leq \rho( B_1^{-1} C_1) < 1.
\end{equation}
Moreover, if $H^{-1} > 0$ and $0 \leq C_2 \leq C_1$ for all components and $C_1 \neq 0$, $C_2 \neq 0$, and $C_2 - C_1 \neq 0$, then
\begin{equation}
    0 < \rho (B_2^{-1} C_2) < \rho( B_1^{-1} C_1) < 1.
\end{equation}
\end{prop}
Following Definition~\ref{def:regular}, and using $H=I-\gamma P$, the Jacobi preconditioner can be seen as a regular splitting $H=\bar B - \bar C$ where $\bar B = \bar H$ and $\bar C = \bar B - H$. Similarly, for standard TD we have that $H=B-C$ where $B = I$ and $C = \gamma P$ forms a valid regular splitting of $H$. With both methods framed in terms of regular splittings, we can now compare their convergence rates using Proposition~\ref{prop:spectral_comp} and setting the learning rate to $1$, i.e, the standard value iteration setting.
\begin{theorem}\label{the:convergence} Let $H=I-\gamma P$ and $\bar H= \diag(H)$, then we have that:
\begin{equation}
    \rho(I - \bar H^{-1} H) \leq \rho(I - H) < 1.
\end{equation}
\begin{proof}
We can rewrite $\rho(I - \bar H^{-1} H)= \rho(\bar H^{-1} \bar C)$ and $\rho(I - H) = \rho(C)$. The rest of the proof follows directly from the properties of regular splittings from Proposition~\ref{prop:spectral_comp}. 
\end{proof}
\end{theorem}
The previous theorem omitted the use of learning rates, in fact, it explicitly assumed a learning rate of $1$. However, it is common to perform a hyperparameter search over the learning rate to obtain the best possible performance. 
Typically, a search over $\alpha$ is performed in order to minimize the spectral radius. To this end, we define the optimal spectral radius, as the minimum over all feasible $\alpha$.
\begin{definition} Given a matrix $H$ only positive real eigenvalues $eig(H)=\{\lambda_1, \lambda_2, ... \} \in \Real^{>0}$ we define the optimal learning rate as:
\begin{equation}
    \alpha^* \deq \argmin_\alpha \rho(I-\alpha H)
\end{equation}
and the resulting optimal spectral radius $\rho^*(I-\alpha H)$:
\begin{equation}
    \rho(I-\alpha^* H) \deq \min_\alpha \rho(I-\alpha H)
\end{equation}
\label{optimal_defs}
\end{definition}
We now focus on finding $\alpha^*$ and the resulting $\rho$ given $H$. First, by standard eigenvalue properties we have the following fact. 
\begin{fact} The eigenvalues of $(I - \alpha H)$  can be rewritten as:
\begin{equation}
    eig(I-\alpha H) = 1 - \alpha eig(H).
\end{equation}
\label{eigen_fact}
\end{fact} 

Next, we rewrite $\rho (I - \alpha H)$ in terms of just the eigenvalues of $A$.
\begin{lemma} For a fixed $\alpha>0$ and a matrix $H$ we have that:
\begin{equation}
    \rho(I - \alpha H) = \max_{\lambda^H} \left \lvert 1 - \alpha \lambda^H \right \rvert .
\end{equation}
\label{max_lambda}
\begin{proof}
The proof comes directly from Fact~\ref{eigen_fact} and the definition of the spectral radius in Def~\ref{spectral_def}.
\end{proof}
\end{lemma}

Next, we rewrite $\rho (I - \alpha H)$ in terms of just the maximum and minimum eigenvalues of $H$.
\begin{lemma} For a fixed $\alpha>0$ and a matrix $H$ with only positive real eigenvalues $eig(H)=\{\lambda_1, \lambda_2, ... \} \in \Real^{>0}$ we have that:
\begin{equation}
    \rho(I - \alpha H) = \max \left \{ \left \lvert 1 - \alpha \lambda_{max}^H \right \rvert , \left \lvert 1 - \alpha \lambda_{min}^H \right \rvert \right \}.
\end{equation}
\label{max_min}
\begin{proof}
The proof follows directly from Prop~\ref{max_lambda} and the assumption that all of the eigenvalues of $H$ are positive.
\end{proof}
\end{lemma}
In words, we have that the spectral radius of $I-\alpha H$ is either a function of the minimum or the maximum eigenvalue of $H$, which we now solve for in the following proposition. Finally, the optimal learning rate $\alpha^*$, for the simple case where $eig(H)\in\Real$, is derived in the following proposition.

\begin{prop}(Corollary 1 from \cite{schoknecht2003td}) For a matrix $H$ with only positive real eigenvalues $eig(H)=\{\lambda_1, \lambda_2, ... \} \in \Real^{>0}$ we have that the $\alpha$ that corresponds to $\min_\alpha \rho(I-\alpha H)$ is:
\begin{equation}
    \alpha^* = \frac{2}{\lambda_{max}^H + \lambda_{min}^H}
\end{equation}
and the resulting spectral radius:
\begin{equation}
    \rho (I - \alpha^* H) = \frac{\lambda_{max}^H - \lambda_{min}^H}{\lambda_{max}^H + \lambda_{min}^H}=\frac{\kappa (H) - 1}{\kappa(H) + 1}.
\end{equation}
\label{optimal_alpha}
where $\kappa=\frac{\lambda_{max}^H}{\lambda_{min}^H}$ is the condition number. 

\begin{proof}
We recall that from lemma~\ref{max_min}, we have that: $$\rho(I - \alpha H)= \max \left \{ \left \lvert 1 - \alpha \lambda_{max}^H \right \rvert , \left \lvert 1 - \alpha \lambda_{min}^H \right \rvert \right \}.$$ 
The $\alpha$ that minimizes this will have the minimal and maximal eigenvalues symmetrically around zero. Therefore $ -(1 +\alpha^* \lambda_{max}^H) =  (1 +\alpha^* \lambda_{min}^H) $ which implies that:
$$
   \alpha^* = \frac{2}{\lambda_{max}^H + \lambda_{min}^H}.
$$
Moreover, plugging $\alpha^*$ into the $\rho(I - \alpha^*H)$ gives us the desired spectral radius.
\end{proof}
\end{prop}
We highlight that from proposition \ref{optimal_alpha}, the \emph{optimal spectral radius} is a monotonically increasing function of the \emph{condition number}, which means that poorly conditioned matrices will induce a slow convergence. As a result, we seek to reduce the condition number of $H$ with the Jacobi preconditioner. By comparing the condition numbers of the Jacobi preconditioned TD and standard TD we can determine which method has better convergence properties and performs best under their respective optimal learning rates. 
\begin{theorem}{(Theorem 10.4.1 \cite{greenbaum1997iterative})} Let $H$, $B_1$, and $B_2$ be symmetric positive-definite matrices satisfying the assumptions of Proposition~\ref{prop:spectral_comp} and suppose that the largest eigenvalue of $B_2^{-1}H \geq 1$. Then the ratios of the largest to smallest eigenvalues of $B_1^{-1}A$ and $B_2^{-1} H$ satisfy:
\begin{equation}
   \frac{\lambda_{max}^{B_1^{-1}H}}{\lambda_{min}^{B_1^{-1}H}} \leq  2 \frac{\lambda_{max}^{B_2^{-1}H}}{\lambda_{min}^{B_2^{-1}H}} .
\end{equation}
\begin{proof}
Since the elements of $B_2^{-1} C_2$ are nonnegative (from Definition~\ref{def:regular}), it follows from Theorem~\ref{the:perron} that its spectral radius corresponds to its largest eigenvalue (which is also simple):
$$
\rho(B_2^{-1} C_2) = \rho(I - B_2^{-1} H) = 1 - \lambda_{min}^{B_2^{-1} H}.
$$

The result $\rho(B_1^{-1} C_1) \leq \rho(B_2^{-1} C_2)$ from Proposition~\ref{prop:spectral_comp} implies that:
$$
1 - \lambda_{min}^{B_1^{-1}H} \leq 1 - \lambda_{min}^{B_2^{-1}H} \text{ and } \lambda_{max}^{B_1^{-1}H} - 1 \leq 1 - \lambda_{min}^{B_2^{-1}H}
$$
or equivalently,
$$
\lambda_{min}^{B_1^{-1}H} \geq \lambda_{min}^{B_2^{-1}H} \text{ and } \lambda_{max}^{B_1^{-1}H} \leq 2 - \lambda_{min}^{B_2^{-1}H}
$$
Dividing the second inequality by the first gives:
\begin{equation}
    \frac{\lambda_{max}^{B_1^{-1}H}}{\lambda_{min}^{B_1^{-1}H}} \leq \frac{\lambda_{max}^{B_2^{-1}H}}{\lambda_{min}^{B_2^{-1}H}} \left( \frac{2 - \lambda_{min}^{B_2^{-1}H}}{\lambda_{max}^{B_2^{-1}H}} \right).
\end{equation}
Since by assumption $\lambda_{max}^{B_1^{-1}H} \geq 1$ and since $\rho(B_2^{-1} C_2) < 1$ implies that $\lambda_{min}^{B_2^{-1}H}>0$, the second factor on the right-hand side is less than $2$.
\end{proof}
\label{the:kappa_comp}
\end{theorem}
\begin{theorem}{(Theorem 10.4.2 \cite{greenbaum1997iterative})}  Let $H$, $B_1$, and $B_2$ be symmetric positive-definite matrices satisfying the assumptions of Proposition~\ref{prop:spectral_comp}, then the assumption in Theorem~\ref{the:kappa_comp} that the largest eigenvalue of $B_2^{-1} H \geq 1$ is satisfied if $H$ and $B_2$ have at least one diagonal element in common.
\begin{proof}
If $H$ and $B_2$ have a diagonal element in common , then the matrix $C_2$ has a zero diagonal element (since $H=B_2 - C_2$). This implies that $B_2^{-1}C_2$ has a nonpositive eigenvalue since the smallest eigenvalue of this matrix satisfies:


\begin{equation*}
\lambda_{\text{min}}^{B_{2}^{-1} C_{1}} = \inf_{v \neq 0} \frac{v^\top C_2 v}{v^\top B_2 v} \leq \frac{u^{\top} C_2 u}{u^\top B_2 u} = 0
\end{equation*}

where $u$ is the vector with a $1$ in the position of the zero diagonal element and $0$s elsewhere. Therefore $B_2^{-1}H = I - B_2^{-1}C_2$ has an eigenvalue greater than or equal to $1$ since:

\begin{align*}
    \lambda_{\text{max}}^{B_{2}^{-1} H} = \sup_{v \neq 0} \bigg\lbrace 1 - \frac{v^\top C_2 v}{v^\top B_2 v} \bigg\rbrace = 1 - \inf_{v \neq 0} \bigg \lbrace \frac{v^\top C_2 v}{v^\top B_2 v} \bigg \rbrace &= 1 - \underbrace{\lambda_{\text{min}}^{B_{2}^{-1} C_{1}}}_{\leq 0} \geq 1
\end{align*}
\end{proof}
\label{the:kappa_comp_extra}
\end{theorem}
Theorems~\ref{the:kappa_comp} and \ref{the:kappa_comp_extra} show that once a pair of regular splittings have been scaled such that $C_2$ has been multiplied by a constant (which does not affect the condition number) that makes $C_2$ have a diagonal element in common with $H$, then the splitting with better convergence rate (in terms of Proposition~\ref{prop:spectral_comp}) has a condition number at worst two times the other. We now apply these results to compare the Jacobi preconditioned system to the original.
\begin{theorem}
\label{the:jacobi_vs_td}
Let $H=\left(I - \gamma P \right )$ and $\bar H=diag\left(I - \gamma P \right)$, then assuming that $H$ is symmetric we have that:
\begin{equation}
    \kappa \left (\bar H^{-1} H \right ) \leq 2 \kappa \left (H \right ).
\end{equation}
\begin{proof}
The result follows directly from Theorems~\ref{the:kappa_comp} and \ref{the:kappa_comp_extra}, and setting $A=H$, $B_1=\bar H$, $B_2=I$.
\end{proof}
\end{theorem}
In words, when $H$ is symmetric, the condition number of the Jacobi preconditioned system is at most a constant factor of $2$ worse than the original system. In practice, we would expect that in the worst, case once a hyperparameter search has been conducted over the learning rate, the Jacobi preconditioner would have similar performance to the original system.

\subsubsection{Extension to n-step returns}
For $n$-step returns we have the following iterative update:
\begin{equation}
    v_{t+1} = v_t - \alpha \left ( H_n v_t -  \sum_{k=0}^{n-1} (\gamma P)^k r \right ),
\end{equation}
where $\alpha$ is the learning rate,  $H_n = \left(I - \gamma^n  P \right)$, and $v_t$ is the estimated value function at time $t$.  By defining the error vector as before, $e_t = v_t - v^*$, we have that $e_{t+1} = (I - \alpha H_n )^{t+1}e_0$.

Applying the Jacobi preconditioner to the $n$-step preconditioned system we get the following iterative formula:
\begin{equation}
    v_{t+1} = v_t - \alpha \bar H_n ^{-1} \left ( H_n v_t -  \sum_{k=0}^{n-1} (\gamma P)^k r \right )
\end{equation}
where following equation~\ref{eq:outer_product_td}, we have $\bar H_n = \diag(H_n) = \diag \left (I - \gamma^n P^n\right)$. We also note that the asymptotic convergence rate of the preconditioned system is $\rho(I - \alpha \bar H_n^{-1} H_n) $.

Following Definition~\ref{def:regular}, and using $H_n = \left(I - \gamma^n  P \right)$, the Jacobi preconditioner can be seen as a regular splitting $H_n=\bar B_n - \bar C_n$ where $\bar B_n = \bar H_n$ and $\bar C = \bar B_n - H_n$. Similarly, for standard $n$-step TD we have that $H_n=B_n-C_n$ where $B_n = I$ and $C_n = I - H_n$ forms a valid regular splitting of $H_n$. Since the requirements for Proposition~\ref{prop:spectral_comp} and Theorem~\ref{the:kappa_comp} apply, analogous results for Theorems~\ref{the:convergence} and \ref{the:jacobi_vs_td} for the $n$-step preconditioned system also hold under the same assumptions. 

\subsubsection{Extension to lambda-returns}
For $\lambda$-returns we have the following iterative update:
\begin{equation}
    v_{t+1} = v_t - \alpha \left ( H_\lambda v_t -  \left(I - \gamma \lambda P \right) ^{-1} r \right ),
\end{equation}
where $\alpha$ is the learning rate,  $H_\lambda = \left(I - \gamma \lambda P \right) ^{-1} \left (I - \gamma P \right )$, and $v_t$ is the estimated value function at time $t$.  By defining the error vector as before, $e_t = v_t - v^*$, we have that $e_{t+1} = (I - \alpha H_\lambda )^{t+1}e_0$.

Applying the Jacobi preconditioner to the $\lambda$ preconditioned system we get the following iterative formula:
\begin{equation}
    v_{t+1} = v_t - \alpha \bar H_\lambda ^{-1} \left ( H_\lambda v_t -  \left(I - \gamma \lambda P \right) ^{-1} r \right )
\end{equation}
where following equation~\ref{eq:outer_product_td}, we have $\bar H_\lambda = \diag(H_\lambda) = \diag \left ( \left(I - \gamma \lambda P \right) ^{-1} \left (I - \gamma P \right ) \right)$. We also note that the asymptotic convergence rate of the preconditioned system is $\rho(I - \alpha \bar H_\lambda^{-1} H_\lambda) $.

Following Definition~\ref{def:regular}, and using $H_\lambda = \left(I - \gamma \lambda P \right) ^{-1} \left (I - \gamma P \right )$, the Jacobi preconditioner can be seen as a regular splitting $H_\lambda=\bar B_\lambda - \bar C_\lambda$ where $\bar B_\lambda = \bar H_\lambda$ and $\bar C = \bar B_\lambda - H_\lambda$. Similarly, for standard TD($\lambda$) we have that $H_\lambda=B_\lambda-C_\lambda$ where $B_\lambda = I$ and $C_\lambda = I - H_\lambda$ forms a valid regular splitting of $H_\lambda$. Since the requirements for Proposition~\ref{prop:spectral_comp} and Theorem~\ref{the:kappa_comp} apply, analogous results for Theorems~\ref{the:convergence} and \ref{the:jacobi_vs_td} for the $\lambda$ preconditioned system also hold under the same assumptions.

\subsection{Extended Experimental Details}
\label{app:experiment}

\subsubsection{Implementation Details}
We perform a random hyperparameter search for TDprop, Adam~\citep{kingma2014adam}, as well as vanilla stochastic gradient descent (SGD), on four Deep RL tasks selected from the Arcade Learning Environment (ALE) \citep{bellemare2013arcade} (we use NoFrameSkip-v4 from OpenAI Gym~\citep{gym}): Beam Rider, Breakout, Qbert, and Space Invaders. Pseudocode is provided for the different optimizers that were used: TDprop in Algorithm~\ref{algo:tdprop}, Adam in Algorithm~\ref{algo:adam}, and SGD in Algorithm~\ref{algo:sgd}. 
We select these four games based on a random sampling from the original DQN benchmark paper~\citep{mnih2013playing}.
We use $n$-step expected SARSA \citep{van2009theoretical}, modifying the A2C implementation of~\citet{pytorchrl}, to train each agent. Specifically, in $16$ parallel threads we sample $5$ transitions using the current policy. We then perform multi-step Expected SARSA updates based on the acquired batch of transitions and repeat the sampling process. Our implementation of Expected SARSA is summarized in Algorithm~\ref{algo:expected_sarsa}. The hyperparameters that were held constant throughout the experiments are summarized in~\ref{tab:fixed_model_hyperparameters_atari}. For the hyperparameter search we sample $50$ random hyperparameter sets from the ranges that are summarized in Table~\ref{tab:hyperparameter_ranges}.  The network architecture is similar to the a2c implementation \citep{pytorchrl} without the layer for the policy, it is summarized in Listing~\ref{code}. Finally, the code repository is located at the following url: \ifdefined\isaccepted \href{https://github.com/joshromoff/tdprop}{https://github.com/joshromoff/tdprop}. \else \href{anonymized}{anonymized}. \fi

\begin{algorithm}[H]
\caption{Synchronous n-step Expected SARSA (per thread)}
\begin{algorithmic}
\small
\Require $\theta$: Parameter vector 
\Require $T=0$: Global shared counter
\State $t\gets 1$ (Initialize thread step counter )
\Repeat
\State $t_0 = t$
\State Get state $s_t$
\Repeat
\State Take action $a_t$ according to the $\epsilon$-greedy policy based on $Q(s_t,a;\theta)$
\State Receive reward $r_t$ and new state $s_{t+1}$
\State $t \gets t + 1$
\State $T \gets T + 1$
\Until  $t - t_0 == n$ 

\For {$i \in \{0,\ldots, n - 1 \}$}
\State Store TD error: $\delta^{\lambda=1}_{t_0:t_0+n - i}= \sum_{k=0}^{k=n-i} \gamma^k r_{i+k} + \gamma^{n-i}\sum_a \pi(s_t, a) Q(s_t,a;\theta) -  Q(s_{t_0+i},a_{t_0+i};\theta)$
\State Store Value: $Q(s_{t_0+i},a_{t_0+i};\theta)$

\EndFor
\State Perform batched update using stored errors and values (from all threads) with Update($\cdot$) from Alg.~\ref{algo:tdprop}, Alg.~\ref{algo:adam}, or Alg.~\ref{algo:sgd}.
\Until $T > T_{max}$
\end{algorithmic}
\label{algo:expected_sarsa}
\end{algorithm}

\begin{algorithm}[H]
\caption{\emph{TDprop}}
\label{algo:tdprop}
\begin{algorithmic}
\Require $\alpha$: Learning rate
\Require $\beta_1,\beta_2 \in [0,1)$: Exponential decay rates for the first and second moment estimates
\Require $\epsilon \in (0,1]$: Damping Hyperparameter
\State $g_0 \gets 0$ (Initialize initial first moment vector) 
\State $z_0 \gets 1$ (Initialize initial second moment vector)
\State $t \gets 0$ (Initialize timestep)
\Function{Update}{$\delta_t$, $v_t$ } (Take as input the TD error $\delta_t$ and the value function $v_t$)
\State $t \gets t + 1$
\State $g_t \gets  \beta_1 \cdot g_{t-1} + (1-\beta_1) \cdot  \delta_t \nabla_\theta v_t$ (Update first moment estimate)
\State $z_t \gets   \beta_2 \cdot z_{t-1} + (1-\beta_2) \cdot ( \nabla_\theta \delta_t \nabla_\theta v_t)^2$ (Update second moment estimate)
\State $\theta_t \gets \theta_{t-1} - \alpha \cdot  g_t / (\sqrt{z_t} + \epsilon)$ (Update parameters)
\EndFunction
\end{algorithmic}
\end{algorithm}

\begin{algorithm}[H]
\caption{\emph{Adam}$^1$}
\label{algo:adam}
\begin{algorithmic}
\Require $\alpha$: Learning rate
\Require $\beta_1,\beta_2 \in [0,1]$: Exponential decay rates for the first and second moment estimates
\Require $\epsilon \in [0,1)$: Damping Hyperparameter
\State $g_0 \gets 0$ (Initialize initial first moment vector)
\State $z_0 \gets 0$ (Initialize initial second moment vector)
\State $t \gets 0$ (Initialize timestep)
\Function{Update}{$\delta_t$, $v_t$ } (Take as input the TD error $\delta_t$ and the value function $v_t$)
\State $t \gets t + 1$
\State $g_t \gets  \beta_1 \cdot g_{t-1} +  (1-\beta_1) \cdot \delta_t \nabla_\theta v_t $ (Update first moment estimate)
\State $z_t \gets  \beta_2 \cdot z_{t-1} + (1-\beta_2) \cdot (\delta_t \nabla_\theta v_t)^2 $ (Update second moment estimate)
\State $\theta_t \gets \theta_{t-1} - \alpha \cdot  g_t / (\sqrt{z_t} + \epsilon)$ (Update parameters)
\EndFunction
\end{algorithmic}
$^{1}$\footnotesize{we omit bias corrections for conciseness.}
\end{algorithm}

\begin{algorithm}[H]
\caption{\emph{SGD}}
\label{algo:sgd}
\begin{algorithmic}
\Require $\alpha$: Learning rate
\Require $\beta_1 \in [0,1)$: Exponential decay rates for the first moment estimates
\State $g_0 \gets 0$ (Initialize initial first moment vector)
\State $t \gets 0$ (Initialize timestep)
\Function{Update}{$\delta_t$, $v_t$ } (Take as input the TD error $\delta_t$ and the value function $v_t$)
\State $t \gets t + 1$
\State $g_t \gets \beta_1\cdot g_{t-1} + (1-\beta_1) \cdot \delta_t \nabla_\theta v_t$ (Update first moment estimate)
\State $\theta_t \gets \theta_{t-1} - \alpha \cdot  g_t$ (Update parameters)
\EndFunction
\end{algorithmic}
\end{algorithm}

\begin{table}[H]
\begin{center}
\begin{tabular}{l@{\hspace{.22cm}}l@{\hspace{.22cm}}l@{\hspace{.22cm}}}
    \toprule
    \textbf{Hyperparameter} & \textbf{Range} & \textbf{Distribution}  \\ \midrule
    
    $(\alpha)$ Learning rate (TDprop and Adam)  & \texttt{[10e-8, 10e-3]} & uniform \\
    $(\alpha)$ Learning rate (SGD) & \texttt{[10e-4, 10e-0]} & uniform \\
    $(\beta_2)$ tracking parameter  & \texttt{[0, 1]} & uniform\\
    $(\epsilon)$ damping parameter  & \texttt{[10e-8, 10e-1]} & uniform\\
    \bottomrule
\end{tabular}
\caption{The ranges used in sampling hyperparameters}
\label{tab:hyperparameter_ranges}
\end{center}
\end{table}

\begin{table}[H]
\begin{center}
\begin{tabular}{l@{\hspace{.22cm}}l@{\hspace{.22cm}}}
\toprule
\textbf{Parameter} & \textbf{Value}  \\
\midrule
Image Width & 84 \\
Image Height & 84 \\
Grayscaling & Yes \\
Action Repetitions & 4 \\
Max-pool over last N action repeat frames & 2 \\
Frame Stacking & 4 \\
End of episode when life lost & Yes \\
Reward Clipping & [-1, 1] \\
Unroll Length ($n$) & 5 \\
Number of Processes & 16 \\
Discount ($\gamma$) & 0.99 \\
Clip global gradient norm & 0.5 \\
$\epsilon$-greedy & 0.01 \\
Number of training steps & 10M \\
      \bottomrule
\end{tabular}
\end{center}
\caption{Hyperparameters for Atari experiments.}
\label{tab:fixed_model_hyperparameters_atari}
\end{table}

\begin{minipage}{\linewidth}
\begin{lstlisting}[language=python,frame=single,caption=Network architecture in pytorch,label=code]
nn.Sequential(
    nn.Conv2d(in_channels=4, out_channels=32, kernel_size=8, stride=4), 
    nn.ReLU(),
    nn.Conv2d(in_channels=32, out_channels=64, kernel_size=4, stride=2), 
    nn.ReLU(), 
    nn.Conv2d(in_channels=64, out_channels=32, kernel_size=3, stride=1),
    nn.ReLU(), 
    nn.Flatten(),
    nn.Linear(in_features=1568, out_features=512),
    nn.ReLU(),
    nn.Linear(in_features=512, out_features=num_actions)
)
\end{lstlisting}
\end{minipage}

\subsubsection{Backpack}
In the mini-batch setting, TDprop needs to compute the required statistic ($- \nabla_{\theta} \delta^\lambda_{t:t+n} \odot \nabla_{\theta} \hat y_t$) for each sample in our mini-batch. Naively, this would increase the computation time by the size of the mini-batch. To alleviate this cost, we parallelize the computation with backpack \citep{dangel2019backpack}, a package for pytorch \citep{paszke2017automatic}.

\subsubsection{Discussion on why expected SARSA?}
While, DQN \citep{mnih2015human}, the Deep version of Q-learning \citep{watkins:thesis89}, is a popular choice for Deep RL, it was not an optimal choice for assessing the effects of Jacobi preconditioning in TD settings, as 1) it requires a target network 2) is off-policy. While TDprop can be extended to the off-policy setting, it is out of the scope of this paper and is left for future work. Another popular choice are Policy Gradient methods \citep{sutton2000policy}, such as A3C \citep{mnih2016asynchronous}, which learn a separate parameterized policy to take actions in the environment. Since actions are not directly chosen from the value function, the impact of an optimizer for the value function is less direct. Instead, we chose to use Expected SARSA \citep{van2009theoretical}, which is an on-policy value based algorithm that has less variance than SARSA with the same amount of bias. We demonstrate that our baseline implementation performs reasonably well in the Atari games tested, given the on-policy nature of the algorithm in Table~\ref{tab:baselines_comparison}. Though DQN, C51, Rainbow, and IQN likely perform better than our implementation of Expected SARSA in 3 out of 4 games, Expected SARSA beats many of these algorithms in Breakout despite not utilizing additional components like target networks, distributional operators, among others. We posit that our implementation of Expected SARSA may be a good starting point for improving the performance of TD algorithms through the \emph{optimizer} rather than through these additional components.
Moreover, SARSA results in our column are taken from the Top 25th percentile of hyperparameters as opposed to multiple random seeds on the most optimal hyperparameters. 
Further improvements to non-optimizer hyperparameters may also improve the performance of our new Expected SARSA baseline implementation in the future.

\begin{table}[H]
    \centering
    \begin{tabular}{|c|c|c|c|c|c|c|c|}
    \hline
         Env& Random & DQN&C51&Rainbow&IQN & A2C & Exp. SARSA (ours)\\
         \hline
         Breakout & 1.2 & 92.2 & 222.4 & 47.9 & 96.3& 303.0 & 123.8 (98.7, 148.1)\\
         BeamRider& 354 & 4064.4&4598.5&5470.7&6211.3&3031.7 & 1753.7 (1621.4, 1864.3)\\
         QBert& 157 & 6836.7 & 9924.5 & 15682.2 & 12496.7& 10065.7 &2968.4 ( 2671.2, 3284.9)\\
         SpaceInvaders& 179 &1178.4 & 1559.1 & 1641.3  & 2558.3&744.5 &586.2 (562.4, 610.1)\\
         \hline
    \end{tabular}
    \caption{Comparison of different baselines published against our implementation of Expected SARSA at 10M timesteps. In many cases SARSA performs significantly worse than off-policy counterparts, yet in Breakout it performs better than most. Off-policy -- DQN~\citep{mnih2013playing}, C51~\citep{bellemare2017distributional}, Rainbow~\citep{hessel2018rainbow}, IQN~\citep{dabney2018implicit}-- results were taken from the Dopamine repository~\citep{castro2018dopamine}. We add (very) rough approximations of A2C performance from \citep{schulman2017proximal}. Note these comparisons may not be fair or exact due to differences in Atari environment used or the approximation of results. Moreover, SARSA results in our column are taken from the Top 25th percentile of hyperparameters as opposed to multiple random seeds on the most optimal hyperparameters. The goal of this table is to demonstrate that performance is roughly in the same range as other baselines, particularly on-policy baselines like A2C.}
    \label{tab:baselines_comparison}
\end{table}

\subsubsection{Discussion on 10M Timesteps}

We elect to evaluate optimizers on the 10 million timesteps of training (equivalent to 40 million frames with frame skipping). This corresponds to roughly 370 hours of game time in Atari. In games with high density rewards, under robust optimization regimes, we believe this should be more than enough time to learn good policies. Additionally, this reduces the energy and carbon costs of experimentation and provides a target for efficient optimization in online TD learning.\citep{henderson2020towards}

\begin{figure*}[!htbp]
    \centering
        \includegraphics[width=0.9\textwidth]{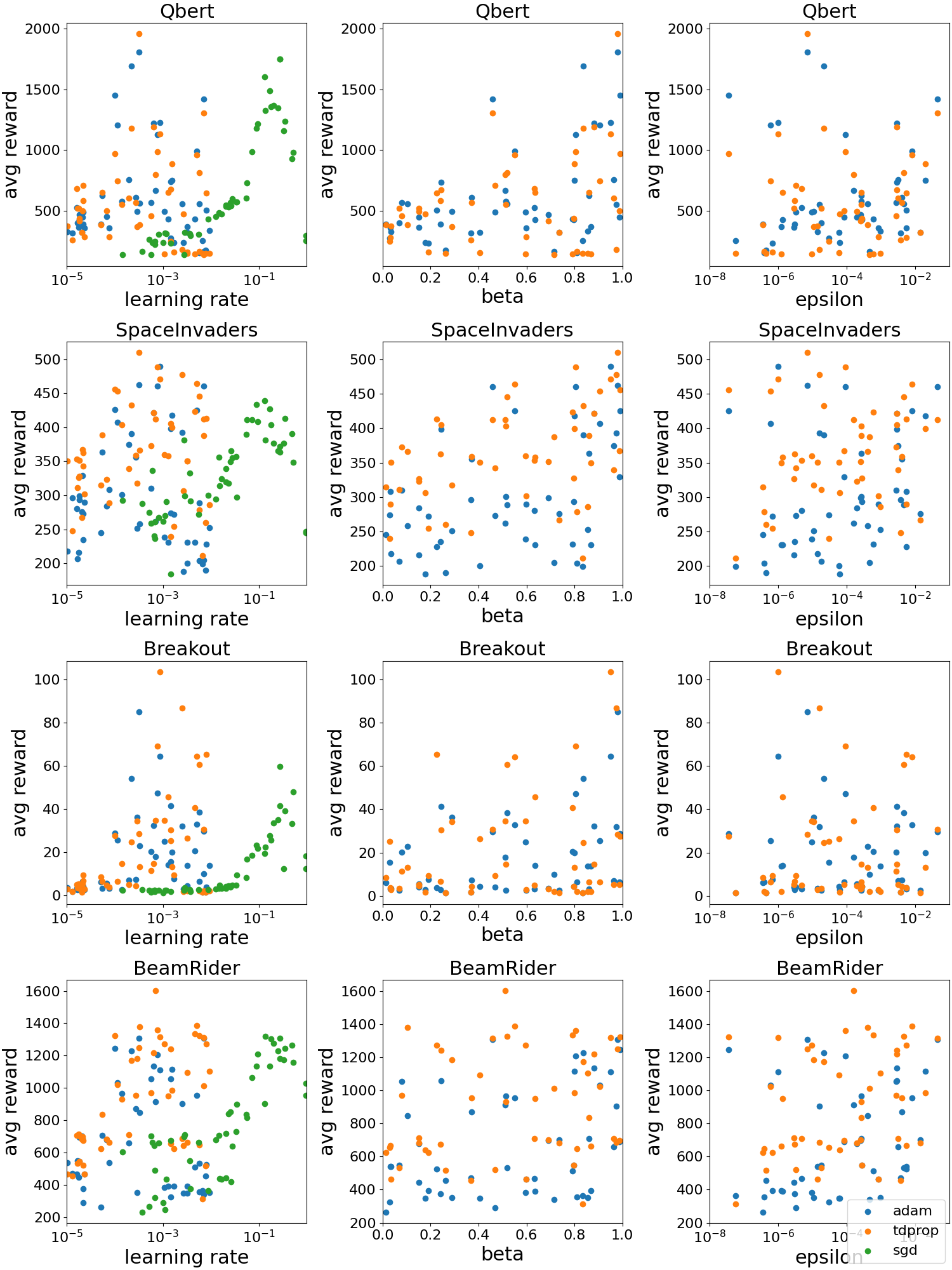}
    
    \caption{Scatter plots of the hyperparameter search on Qbert,  Breakout, Space Invaders, and Beam Rider. The y-axis represents the average undiscounted return per episode over $10$ million training steps. }
    \label{fig:appresults1}
\end{figure*}

\begin{figure*}[!htbp]
    \centering
        \includegraphics[width=0.9\textwidth]{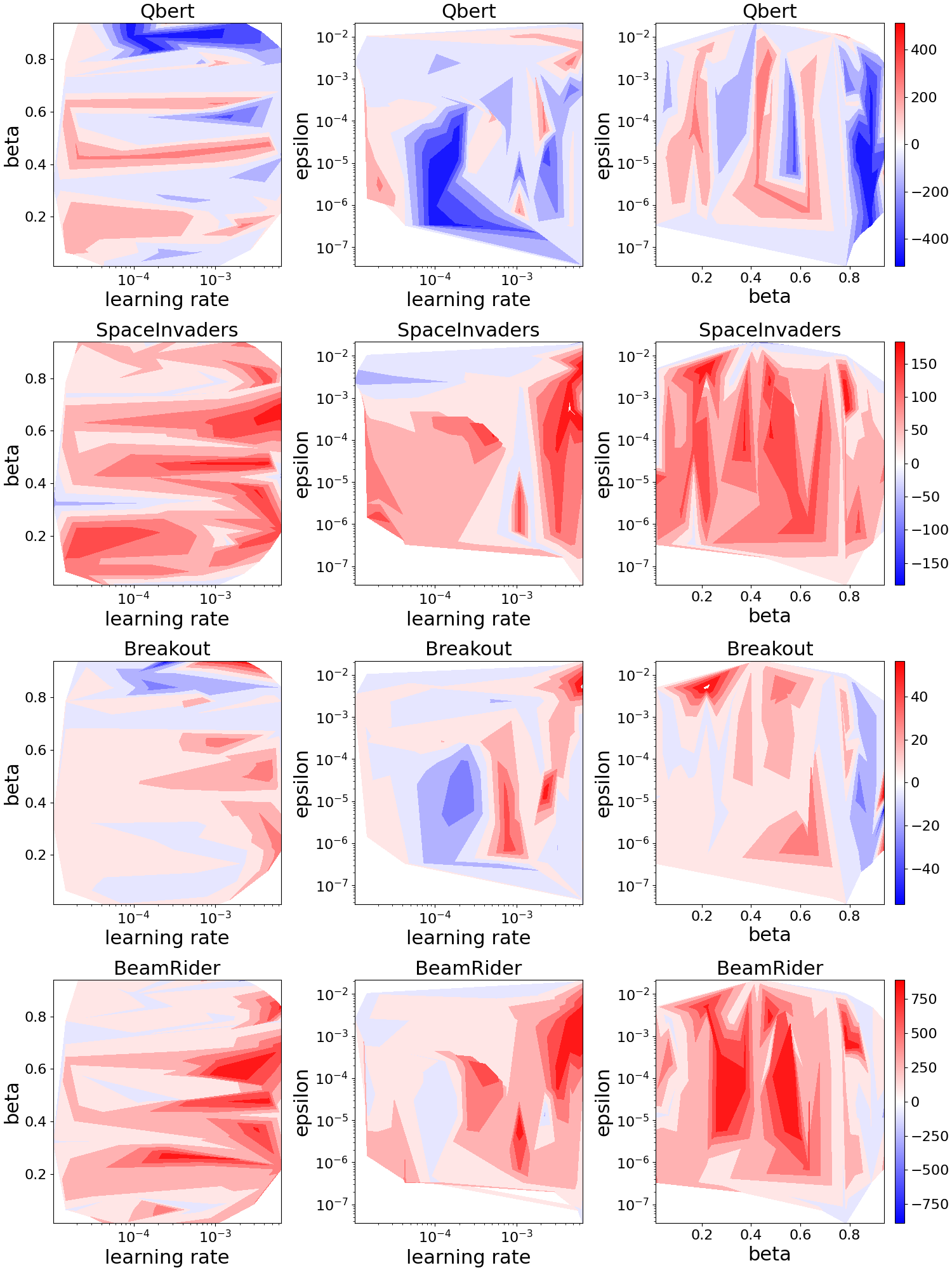}
    
    \caption{Histograms of the hyperparameter search on Qbert,  Breakout, Space Invaders, and Beam Rider.}
    \label{fig:appresults2}

\end{figure*}

\subsubsection{Adam $\beta_1=0$}

We note that we use $\beta_1=0$ for Adam throughout this work. We do so to emphasize the investigation on the per-parameter learning rates rather than the gradient smoothing role that $\beta_1$ plays. We also note that per \citet{kingma2014adam}, if $\beta_1=0$ Adam is similar to Adagrad~\citep{duchi2011adaptive}. \citet{sun2020adaptive} investigate convergence rates for TD using Adagrad in linear settings, they find similar worst case bounds on performance as non-adaptive TD. Those results may explain some of the parity in performance that we find here. We also note that $\beta_1=0$ has been used to great success empirically \citep{espeholt2018impala}. That being said, we caveat that results may change if $\beta_1$ is used in both TDprop and Adam simultaneously. From now on all references to $\beta$ refer to the $\beta_2$ parameter.

\subsection{Extended Analysis}
\label{app:analysis}

\begin{figure}[H]
    \centering
    \includegraphics[width=.48\textwidth]{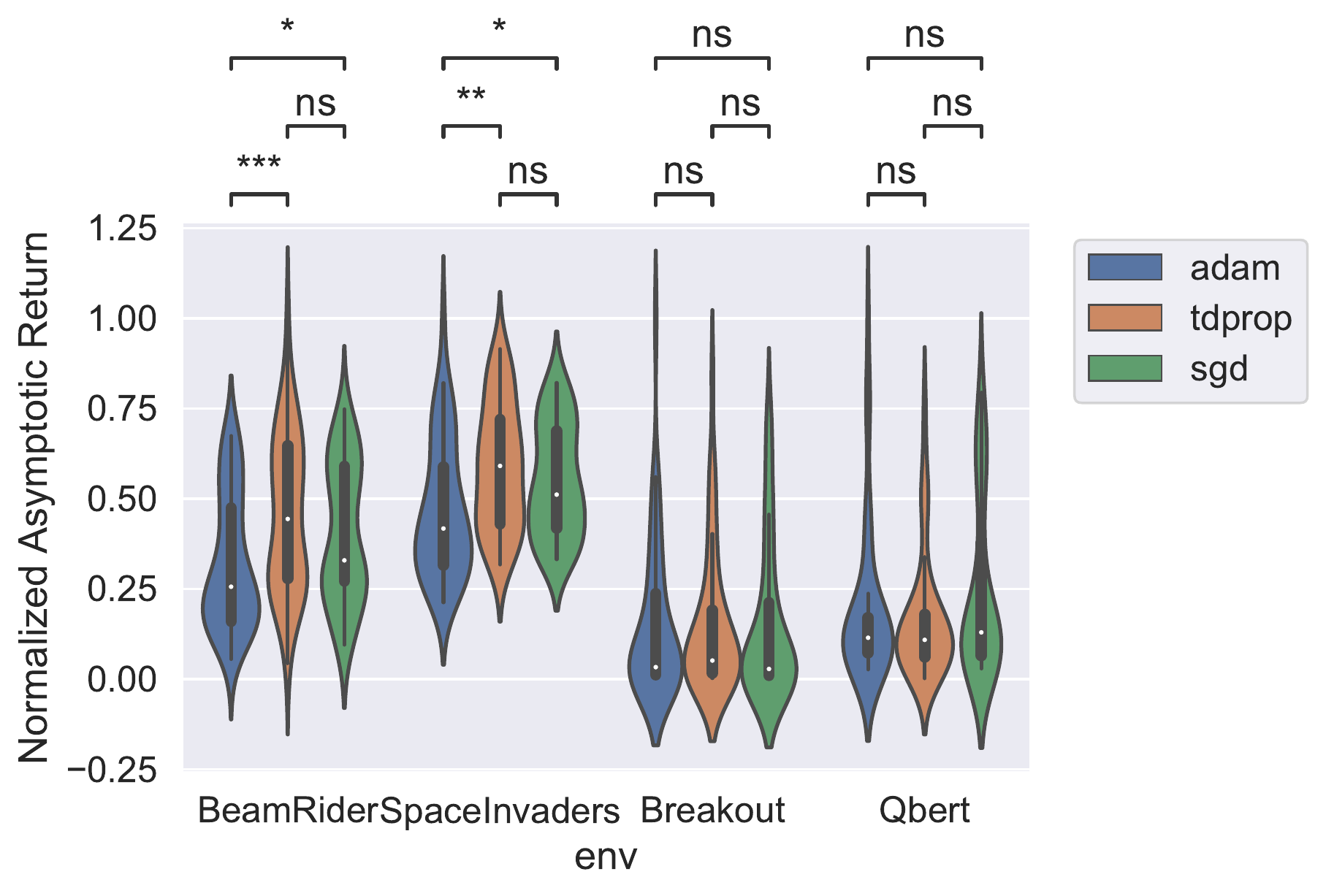}
    \includegraphics[width=.48\textwidth]{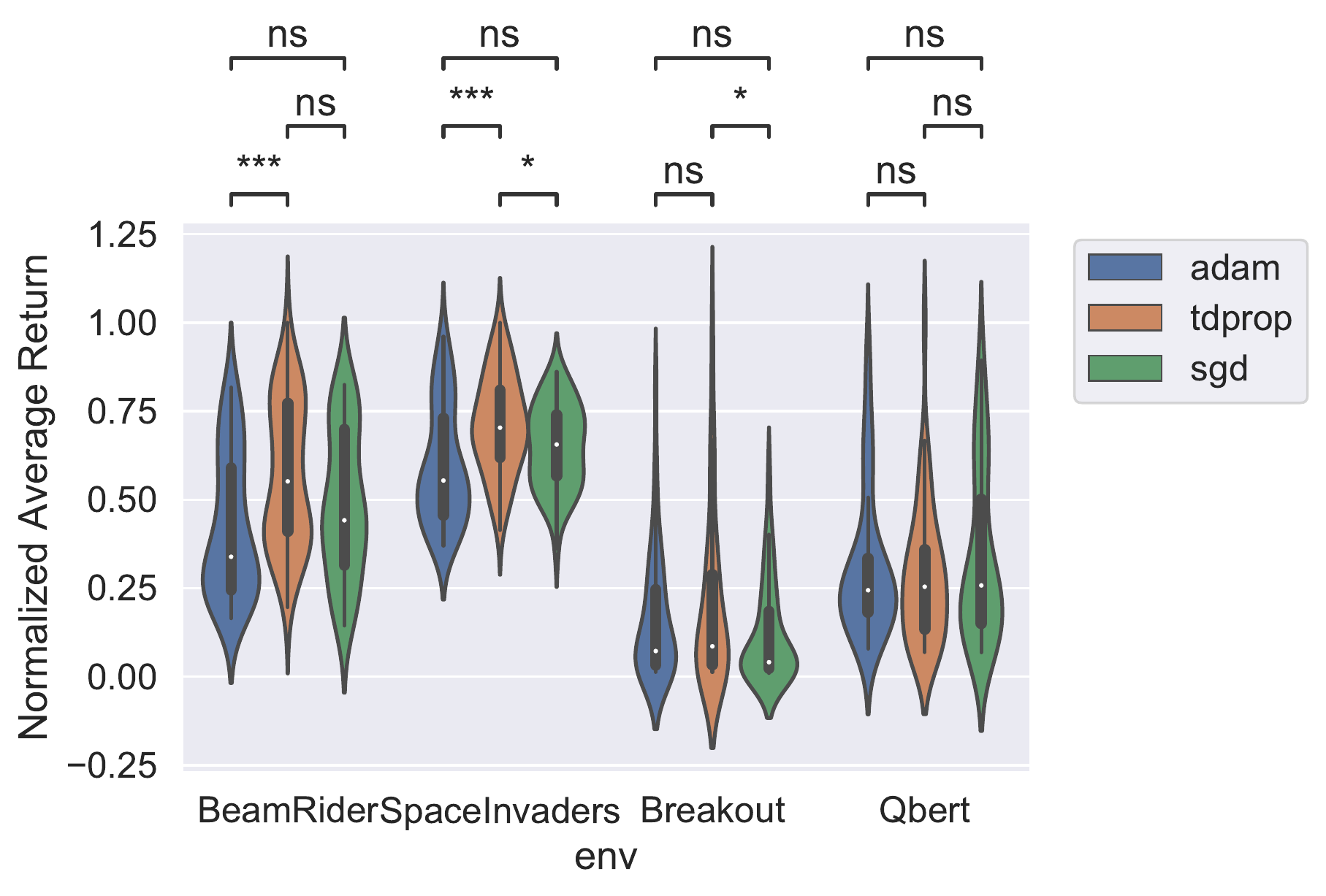}
    \caption{The normalized asymptotic returns of all hyperparameter configurations attempted (left) and the normalized average returns of all hyperparameter configurations attempted (right). Normalization is performed by taking the maximum value for the game and dividing all results by this value. Significance tests are done using Welch's t-test, per recommendations from \citet{colas2019hitchhiker,henderson2018deep}. P-value annotation legend is as follows. ns: $0.05 < p <= 1$; *: $0.01 < p <= 0.05$; **: $0.001 < p <= 0.01$; ***: $0.0001 < p <=  0.001$; ****: $p <= 0.0001$.}
    \label{fig:violinplots1}
\end{figure}
\begin{figure}[H]
    \centering
    \includegraphics[width=.48\textwidth]{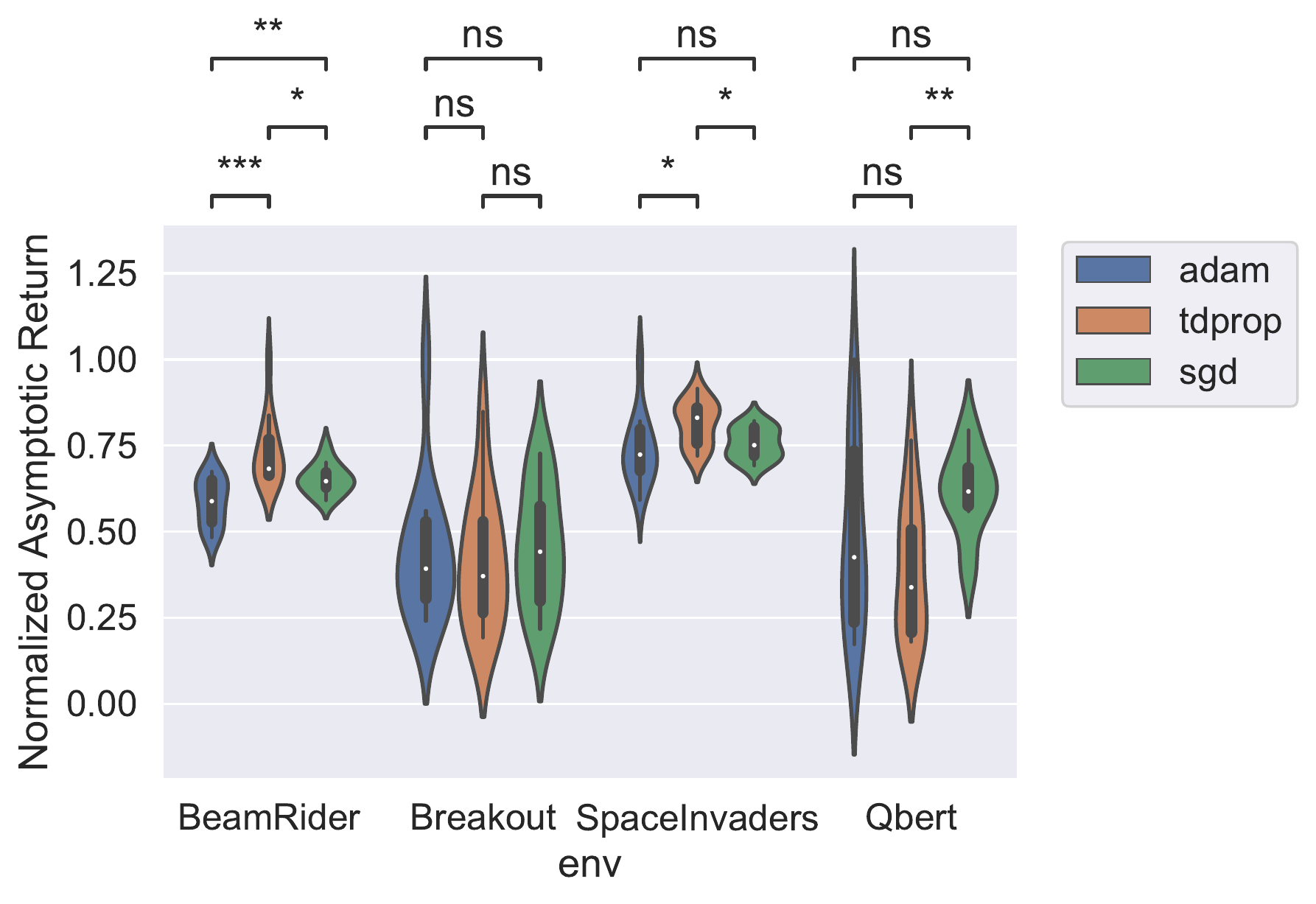}
    \includegraphics[width=.48\textwidth]{images/top_25_avg_plot.pdf}
    \caption{The normalized asymptotic returns of all hyperparameter configurations  within the top 25th percentile (left) and the normalized average returns of all hyperparameter configurations within the top 25th percentile (right). Normalization is performed by taking the maximum value for the game and dividing all results by this value. Significance tests are done using Welch's t-test, per recommendations from \citet{colas2019hitchhiker,henderson2018deep}. P-value annotation legend is as follows. ns: $0.05 < p <= 1$; *: $0.01 < p <= 0.05$; **: $0.001 < p <= 0.01$; ***: $0.0001 < p <=  0.001$; ****: $p <= 0.0001$.}
    \label{fig:violinplots2}
\end{figure}

Figures~\ref{fig:appresults1},~\ref{fig:appresults2},~\ref{fig:violinplots1}, and~\ref{fig:violinplots2} show our results across the distribution of hyperparameters. We also summarize results in Table~\ref{tab:results10mas} and~\ref{tab:results10mav} with bootstrap confidence intervals.

In the Regression Tables concluding the Appendix, we show regressions on the hyperparameters and their effect on average return across multiple games. Interestingly for TDprop, the $\beta$ parameter seems to play a less significant role in BeamRider and Breakout than in other games and learning rates on their own play a lesser role as well.
This is contrary to Adam, where it plays a significant role across all games. 
Since the $\beta$ parameter plays a role in smoothing the approximation of the inverse Hessian diagonal.
We suspect that if the data distribution is relatively stationary and not not noisy (e.g., the batch size is large enough) then it may not matter what $\beta$ you choose. Alternatively, a large $\beta$ may be necessary for noisy settings or settings with small batch sizes.
There may be an interplay with the accuracy of the second order approximation. However, for all these hypotheses we do not have enough data to make conclusions as for the causal reason for the reduced role of $\beta$ in these games.



Interestingly, TDProp $R^2$ are much lower than Adam. This might be because TDprop is can have more sensitive non-linear interactions between hyperparameters (see Figure~\ref{fig:appresults1}). For example, the denominator can go to zero relying on the epsilon to prevent divergence. However, we do not have enough data to back this hypothesis. Additionally, because of the poor linear fit, we note that these analyses may deviate if a non-linear approach were taken to analyzing the data -- and thus the regression analysis for TDProp should be only viewed as a rough guide rather than an absolute truth.

We also find that in certain tasks SGD prefers a considerably larger learning rate than both aforementioned methods, roughly two orders of magnitude larger at approximately $>10^{-0.5}$ for Qbert, Breakout, and Beam Rider. 
This discrepancy is not seen for both TDprop and Adam, which both tend to have values close to $10^{-3}$ as the optimal choice across tasks.

We compare the effect of each hyperparameter for TDprop and Adam in Figure~\ref{fig:appresults2}. Specifically, we measure the difference in performance between TDprop and Adam (TDprop - Adam) across tasks and hyperparameters. What we find is that in most tasks (except for Qbert) TDprop has a better overall coverage of the hyperparameter space, suggesting it improves stability in the non-convex regime. 

This is also reflected in Tables~\ref{tab:results10mav} and~\ref{tab:results10mas}, as well as Figures~\ref{fig:violinplots1} and~\ref{fig:violinplots2}. TDProp performs significantly better in the top 25th percentile of hyperparameter results some of the time (as seen in Figure~\ref{fig:violinplots2}) (for example, it is best or tied for best in BeamRider, Breakout, and SpaceInvaders on both average return and asymptotic return, but does significantly worse than SGD in Qbert. Interestingly,  SGD outperforms both Adam and TDprop in some cases and ties in other cases.

\begin{table}[H]
    \centering
    \begin{tabular}{|c|c|c|c|}
        \hline
        \multicolumn{4}{c}{All Hyperparameter Samples}\\
        \hline
         Game & SGD & Adam & TDprop \\
         \hline
BeamRider&\textit{778.6 (686.4, 867.7)} $\dagger$& \textit{673.9 (584.4, 760.4)} $\dagger$ & \textbf{907.7 (815.3, 995.3)}\\
Breakout&12.2 (7.9, 16.0) & 16.9 (11.7, 21.7) & 20.2 (13.3, 26.5)\\
SpaceInvaders&\textit{330.6 (314.3, 347.5)} $\dagger$ & 302.3 (278.3, 325.1) & \textbf{362.3 (343.1, 381.3)}\\
Qbert&654.3 (519.5, 781.7) & 599.2 (483.4, 703.9) & 552.3 (444.0, 651.0)\\
        \hline
        \multicolumn{4}{c}{Top 25\%}\\
        \hline
BeamRider&\textit{1226.2 (1192.4, 1259.7)} $\dagger$ & 1131.8 (1069.3, 1192.5) & \textbf{1336.2 (1282.2, 1377.5)} \\
Breakout&33.0 (26.5, 38.9) & 42.1 (32.5, 50.0) & 53.9 (41.1, 65.1) $^*$ \\
SpaceInvaders&\textit{404.9 (394.3, 415.5)} & \textit{425.0 (407.1, 442.3)}$\dagger$ & \textbf{451.4 (435.7, 467.0) } \\
Qbert&1366.8 (1243.9, 1483.3)$^*$ & 1157.8 (956.8, 1351.3) & 1048.5 (860.6, 1199.5) \\
        \hline
    \end{tabular}
    \caption{For up to 10M timesteps. Average return with bootstrap confidence intervals in parentheses. Bolded text indicates best based on bootstrap significance test. $\dagger$ indicates runner up by significance testing. If multiple values fall into a tier, denote them by the same marker. For the top 25\% of TDprop vs SGD on QBert, the only significant comparison is against SGD (TDProp is significantly worse than SGD, but Adam is not significantly better (or worse) than TDprop or SGD. Conversely for SGD and TDprop on Breakout. This is indicated by $^*$}
    \label{tab:results10mav}
\end{table}

\begin{table}[H]
    \centering
    \begin{tabular}{|c|c|c|c|}
        \hline
        \multicolumn{4}{c}{All Hyperparameter Samples}\\
        \hline
         Game & SGD & Adam & TDprop \\
         \hline
BeamRider&\textbf{963.7 (834.4, 1085.4)} & \textit{765.6 (643.1, 884.1)}$\dagger$ & \textbf{1091.5 (948.0, 1230.9)}\\
Breakout&39.4 (23.6, 54.1) & 40.3 (23.8, 54.8) & 39.9 (25.1, 53.4)\\
SpaceInvaders&\textbf{394.1 (363.1, 424.3) }& \textit{336.6 (298.2, 373.0)}$\dagger$ & \textbf{420.8 (386.8, 455.0)}\\
Qbert&1144.4 (818.0, 1451.5)$^*$ & 930.4 (627.0, 1195.6) & 792.9 (550.2, 1002.6)\\
        \hline
        \multicolumn{4}{c}{Top 25\%}\\
        \hline
BeamRider&\textit{1559.9 (1503.6, 1613.6)} $\dagger$ & 1402.9 (1320.7, 1489.3) & \textbf{1753.7 (1621.4, 1864.3) }\\
Breakout&123.8 (98.7, 148.1) & 119.0 (87.2, 143.5) & 112.2 (83.5, 137.3) \\
SpaceInvaders&\textit{543.8 (527.2, 560.7)} $\dagger$ & \textit{530.2 (488.5, 564.8)}$\dagger$ & \textbf{586.2 (562.4, 610.1)} \\
Qbert&2968.4 (2671.2, 3284.9)$^*$ & 2284.4 (1603.6, 2915.3) & 1869.4 (1376.5, 2337.4) \\
        \hline
    \end{tabular}
    \caption{For up to 10M timesteps. Asymptotic return with bootstrap confidence intervals in parentheses. Bolded text indicates best based on bootstrap significance test. $\dagger$ indicates runner up by significance testing. If multiple values fall into a tier, denote them by the same marker. For the top 25\% of TDprop vs SGD on QBert, the only significant comparison is against SGD (TDProp is significantly worse than SGD, but Adam is not significantly better (or worse) than TDprop or SGD. This is indicated by $^*$}
    \label{tab:results10mas}
\end{table}

\begin{table}[!htbp] \centering 
  \caption{TDProp regression for 10M timesteps with scaling and interaction terms.} 
\begin{tabular}{@{\extracolsep{5pt}}lD{.}{.}{-3} D{.}{.}{-3} D{.}{.}{-3} D{.}{.}{-3} } 
\\[-1.8ex]\hline 
\hline \\[-1.8ex] 
 & \multicolumn{4}{c}{\textit{Dependent variable:}} \\ 
\cline{2-5} 
\\[-1.8ex] & \multicolumn{4}{c}{scale(avg\_return)} \\ 
\\[-1.8ex] & \multicolumn{1}{c}{BeamRider} & \multicolumn{1}{c}{Breakout} & \multicolumn{1}{c}{SpaceInvaders} & \multicolumn{1}{c}{QBert}\\ 
\hline \\[-1.8ex] 
 scale(lr) & 0.374 & 0.607^{*} & 0.248 & 0.154 \\ 
  & (0.362) & (0.358) & (0.320) & (0.299) \\ 
  & & & & \\ 
 scale(beta) & 0.320^{*} & 0.378^{**} & 0.608^{***} & 0.484^{***} \\ 
  & (0.178) & (0.177) & (0.158) & (0.147) \\ 
  & & & & \\ 
 scale(epsilon) & 0.171 & 0.112 & -0.186 & -0.405 \\ 
  & (1.144) & (1.133) & (1.012) & (0.944) \\ 
  & & & & \\ 

 scale(lr \textasteriskcentered  epsilon) & 0.222 & 0.225 & 0.508 & 0.650 \\ 
  & (0.518) & (0.513) & (0.458) & (0.427) \\ 
  & & & & \\ 
 scale(lr \textasteriskcentered  beta) & -0.358 & -0.585 & -0.541^{*} & -0.648^{**} \\ 
  & (0.361) & (0.358) & (0.320) & (0.298) \\ 
  & & & & \\ 
 scale(epsilon \textasteriskcentered  beta) & -0.223 & -0.339 & -0.158 & 0.199 \\ 
  & (0.787) & (0.779) & (0.696) & (0.649) \\ 
  & & & & \\ 
 Constant & -0.000 & -0.000 & 0.000 & -0.000 \\ 
  & (0.140) & (0.139) & (0.124) & (0.116) \\ 
  & & & & \\ 
\hline \\[-1.8ex] 
Observations & \multicolumn{1}{c}{50} & \multicolumn{1}{c}{50} & \multicolumn{1}{c}{50} & \multicolumn{1}{c}{50} \\ 
R$^{2}$ & \multicolumn{1}{c}{0.139} & \multicolumn{1}{c}{0.155} & \multicolumn{1}{c}{0.326} & \multicolumn{1}{c}{0.414} \\ 
Adjusted R$^{2}$ & \multicolumn{1}{c}{0.018} & \multicolumn{1}{c}{0.037} & \multicolumn{1}{c}{0.232} & \multicolumn{1}{c}{0.332} \\ 
Residual Std. Error (df = 43) & \multicolumn{1}{c}{0.991} & \multicolumn{1}{c}{0.981} & \multicolumn{1}{c}{0.876} & \multicolumn{1}{c}{0.817} \\ 
F Statistic (df = 6; 43) & \multicolumn{1}{c}{1.153} & \multicolumn{1}{c}{1.317} & \multicolumn{1}{c}{3.466$^{***}$} & \multicolumn{1}{c}{5.062$^{***}$} \\ 
\hline 
\hline \\[-1.8ex] 
\textit{Note:}  & \multicolumn{4}{r}{$^{*}$p$<$0.1; $^{**}$p$<$0.05; $^{***}$p$<$0.01} \\ 
\end{tabular} 
\end{table}

\begin{table}[!htbp] \centering 
  \caption{TDProp regression for 10m steps without interaction terms, but with scaling.} 
\begin{tabular}{@{\extracolsep{5pt}}lD{.}{.}{-3} D{.}{.}{-3} D{.}{.}{-3} D{.}{.}{-3} } 
\\[-1.8ex]\hline 
\hline \\[-1.8ex] 
 & \multicolumn{4}{c}{\textit{Dependent variable:}} \\ 
\cline{2-5} 
\\[-1.8ex] & \multicolumn{4}{c}{scale(avg\_return)} \\ 
\\[-1.8ex] & \multicolumn{1}{c}{BeamRider} & \multicolumn{1}{c}{Breakout} & \multicolumn{1}{c}{SpaceInvaders} & \multicolumn{1}{c}{QBert}\\ 
\hline \\[-1.8ex] 
 scale(lr) & 0.103 & 0.143 & -0.147 & -0.335^{**} \\ 
  & (0.146) & (0.149) & (0.137) & (0.130) \\ 
  & & & & \\ 
 scale(beta) & 0.202 & 0.194 & 0.441^{***} & 0.325^{**} \\ 
  & (0.141) & (0.143) & (0.131) & (0.125) \\ 
  & & & & \\ 
 scale(epsilon) & 0.175 & 0.024 & 0.143 & 0.402^{***} \\ 
  & (0.146) & (0.148) & (0.136) & (0.130) \\ 
  & & & & \\ 
 Constant & -0.000 & -0.000 & 0.000 & -0.000 \\ 
  & (0.139) & (0.141) & (0.130) & (0.123) \\ 
  & & & & \\ 
\hline \\[-1.8ex] 
Observations & \multicolumn{1}{c}{50} & \multicolumn{1}{c}{50} & \multicolumn{1}{c}{50} & \multicolumn{1}{c}{50} \\ 
R$^{2}$ & \multicolumn{1}{c}{0.095} & \multicolumn{1}{c}{0.066} & \multicolumn{1}{c}{0.213} & \multicolumn{1}{c}{0.285} \\ 
Adjusted R$^{2}$ & \multicolumn{1}{c}{0.036} & \multicolumn{1}{c}{0.005} & \multicolumn{1}{c}{0.161} & \multicolumn{1}{c}{0.238} \\ 
Residual Std. Error (df = 46) & \multicolumn{1}{c}{0.982} & \multicolumn{1}{c}{0.998} & \multicolumn{1}{c}{0.916} & \multicolumn{1}{c}{0.873} \\ 
F Statistic (df = 3; 46) & \multicolumn{1}{c}{1.612} & \multicolumn{1}{c}{1.077} & \multicolumn{1}{c}{4.144$^{**}$} & \multicolumn{1}{c}{6.109$^{***}$} \\ 
\hline 
\hline \\[-1.8ex] 
\textit{Note:}  & \multicolumn{4}{r}{$^{*}$p$<$0.1; $^{**}$p$<$0.05; $^{***}$p$<$0.01} \\ 
\end{tabular} 
\end{table} 

\begin{table}[!htbp] \centering 
  \caption{Adam regression to 10m timesteps with interaction terms and scaling.} 
\begin{tabular}{@{\extracolsep{5pt}}lD{.}{.}{-3} D{.}{.}{-3} D{.}{.}{-3} D{.}{.}{-3} } 
\\[-1.8ex]\hline 
\hline \\[-1.8ex] 
 & \multicolumn{4}{c}{\textit{Dependent variable:}} \\ 
\cline{2-5} 
\\[-1.8ex] & \multicolumn{4}{c}{scale(avg\_return)} \\ 
\\[-1.8ex] & \multicolumn{1}{c}{BeamRider} & \multicolumn{1}{c}{Breakout} & \multicolumn{1}{c}{SpaceInvaders} & \multicolumn{1}{c}{QBert}\\ 
\hline \\[-1.8ex] 
 scale(lr) & -0.001 & -0.001 & -0.135 & 0.009 \\ 
  & (0.293) & (0.352) & (0.259) & (0.277) \\ 
  & & & & \\ 
 scale(beta) & 0.564^{***} & 0.456^{**} & 0.675^{***} & 0.646^{***} \\ 
  & (0.144) & (0.173) & (0.128) & (0.137) \\ 
  & & & & \\ 
 scale(epsilon) & 1.324 & 0.105 & 1.465^{*} & 0.235 \\ 
  & (0.925) & (1.112) & (0.819) & (0.877) \\ 
  & & & & \\ 

 scale(lr \textasteriskcentered  epsilon) & -0.188 & 0.309 & -0.193 & 0.518 \\ 
  & (0.419) & (0.504) & (0.371) & (0.397) \\ 
  & & & & \\ 
 scale(lr \textasteriskcentered  beta) & -0.431 & -0.174 & -0.344 & -0.485^{*} \\ 
  & (0.292) & (0.351) & (0.259) & (0.277) \\ 
  & & & & \\ 
 scale(epsilon \textasteriskcentered  beta) & -0.753 & -0.291 & -0.880 & -0.369 \\ 
  & (0.636) & (0.765) & (0.563) & (0.603) \\ 
  & & & & \\ 
 Constant & -0.000 & 0.000 & 0.000 & -0.000 \\ 
  & (0.113) & (0.136) & (0.100) & (0.107) \\ 
  & & & & \\ 
\hline \\[-1.8ex] 
Observations & \multicolumn{1}{c}{50} & \multicolumn{1}{c}{50} & \multicolumn{1}{c}{50} & \multicolumn{1}{c}{50} \\ 
R$^{2}$ & \multicolumn{1}{c}{0.437} & \multicolumn{1}{c}{0.186} & \multicolumn{1}{c}{0.559} & \multicolumn{1}{c}{0.494} \\ 
Adjusted R$^{2}$ & \multicolumn{1}{c}{0.358} & \multicolumn{1}{c}{0.072} & \multicolumn{1}{c}{0.497} & \multicolumn{1}{c}{0.423} \\ 
Residual Std. Error (df = 43) & \multicolumn{1}{c}{0.801} & \multicolumn{1}{c}{0.963} & \multicolumn{1}{c}{0.709} & \multicolumn{1}{c}{0.759} \\ 
F Statistic (df = 6; 43) & \multicolumn{1}{c}{5.553$^{***}$} & \multicolumn{1}{c}{1.638} & \multicolumn{1}{c}{9.079$^{***}$} & \multicolumn{1}{c}{6.994$^{***}$} \\ 
\hline 
\hline \\[-1.8ex] 
\textit{Note:}  & \multicolumn{4}{r}{$^{*}$p$<$0.1; $^{**}$p$<$0.05; $^{***}$p$<$0.01} \\ 
\end{tabular} 
\end{table} 

\begin{table}[!htbp] \centering 
  \caption{Adam regression with scaling and without interaction terms to 10m steps.} 
\begin{tabular}{@{\extracolsep{5pt}}lD{.}{.}{-3} D{.}{.}{-3} D{.}{.}{-3} D{.}{.}{-3} } 
\\[-1.8ex]\hline 
\hline \\[-1.8ex] 
 & \multicolumn{4}{c}{\textit{Dependent variable:}} \\ 
\cline{2-5} 
\\[-1.8ex] & \multicolumn{4}{c}{scale(avg\_return)} \\ 
\\[-1.8ex] & \multicolumn{1}{c}{BeamRider} & \multicolumn{1}{c}{Breakout} & \multicolumn{1}{c}{SpaceInvaders} & \multicolumn{1}{c}{QBert}\\ 
\hline \\[-1.8ex] 
 scale(lr) & -0.367^{***} & -0.092 & -0.416^{***} & -0.322^{**} \\ 
  & (0.121) & (0.142) & (0.108) & (0.123) \\ 
  & & & & \\ 
 scale(beta) & 0.390^{***} & 0.369^{***} & 0.507^{***} & 0.469^{***} \\ 
  & (0.117) & (0.137) & (0.104) & (0.118) \\ 
  & & & & \\ 
 scale(epsilon) & 0.464^{***} & 0.102 & 0.469^{***} & 0.362^{***} \\ 
  & (0.121) & (0.142) & (0.108) & (0.122) \\ 
  & & & & \\ 
 Constant & 0.000 & 0.000 & 0.000 & -0.000 \\ 
  & (0.115) & (0.135) & (0.103) & (0.117) \\ 
  & & & & \\ 
\hline \\[-1.8ex] 
Observations & \multicolumn{1}{c}{50} & \multicolumn{1}{c}{50} & \multicolumn{1}{c}{50} & \multicolumn{1}{c}{50} \\ 
R$^{2}$ & \multicolumn{1}{c}{0.381} & \multicolumn{1}{c}{0.143} & \multicolumn{1}{c}{0.503} & \multicolumn{1}{c}{0.362} \\ 
Adjusted R$^{2}$ & \multicolumn{1}{c}{0.341} & \multicolumn{1}{c}{0.087} & \multicolumn{1}{c}{0.471} & \multicolumn{1}{c}{0.320} \\ 
Residual Std. Error (df = 46) & \multicolumn{1}{c}{0.812} & \multicolumn{1}{c}{0.955} & \multicolumn{1}{c}{0.727} & \multicolumn{1}{c}{0.824} \\ 
F Statistic (df = 3; 46) & \multicolumn{1}{c}{9.451$^{***}$} & \multicolumn{1}{c}{2.558$^{*}$} & \multicolumn{1}{c}{15.548$^{***}$} & \multicolumn{1}{c}{8.698$^{***}$} \\ 
\hline 
\hline \\[-1.8ex] 
\textit{Note:}  & \multicolumn{4}{r}{$^{*}$p$<$0.1; $^{**}$p$<$0.05; $^{***}$p$<$0.01} \\ 
\end{tabular} 
\end{table} 

\begin{table}[!htbp] \centering 
  \caption{Adam no scaling no interaction terms} 
\begin{tabular}{@{\extracolsep{5pt}}lD{.}{.}{-3} D{.}{.}{-3} D{.}{.}{-3} D{.}{.}{-3} } 
\\[-1.8ex]\hline 
\hline \\[-1.8ex] 
 & \multicolumn{4}{c}{\textit{Dependent variable:}} \\ 
\cline{2-5} 
\\[-1.8ex] & \multicolumn{4}{c}{(avg\_return)} \\ 
\\[-1.8ex] & \multicolumn{1}{c}{BeamRider} & \multicolumn{1}{c}{Breakout} & \multicolumn{1}{c}{SpaceInvaders} & \multicolumn{1}{c}{QBert}\\ 
\hline \\[-1.8ex] 
 lr & -43,966.180^{***} & -628.987 & -13,221.510^{***} & -47,934.430^{**} \\ 
  & (14,518.190) & (974.753) & (3,445.708) & (18,289.160) \\ 
  & & & & \\ 
 beta & 388.670^{***} & 20.946^{***} & 133.800^{***} & 579.182^{***} \\ 
  & (116.078) & (7.794) & (27.550) & (146.229) \\ 
  & & & & \\ 
 epsilon & 20,514.360^{***} & 256.975 & 5,501.973^{***} & 19,877.590^{***} \\ 
  & (5,332.311) & (358.012) & (1,265.556) & (6,717.331) \\ 
  & & & & \\ 
 Constant & 494.060^{***} & 6.145 & 240.792^{***} & 325.777^{***} \\ 
  & (75.638) & (5.078) & (17.952) & (95.284) \\ 
  & & & & \\ 
\hline \\[-1.8ex] 
Observations & \multicolumn{1}{c}{50} & \multicolumn{1}{c}{50} & \multicolumn{1}{c}{50} & \multicolumn{1}{c}{50} \\ 
R$^{2}$ & \multicolumn{1}{c}{0.381} & \multicolumn{1}{c}{0.143} & \multicolumn{1}{c}{0.503} & \multicolumn{1}{c}{0.362} \\ 
Adjusted R$^{2}$ & \multicolumn{1}{c}{0.341} & \multicolumn{1}{c}{0.087} & \multicolumn{1}{c}{0.471} & \multicolumn{1}{c}{0.320} \\ 
Residual Std. Error (df = 46) & \multicolumn{1}{c}{259.812} & \multicolumn{1}{c}{17.444} & \multicolumn{1}{c}{61.663} & \multicolumn{1}{c}{327.296} \\ 
F Statistic (df = 3; 46) & \multicolumn{1}{c}{9.451$^{***}$} & \multicolumn{1}{c}{2.558$^{*}$} & \multicolumn{1}{c}{15.548$^{***}$} & \multicolumn{1}{c}{8.698$^{***}$} \\ 
\hline 
\hline \\[-1.8ex] 
\textit{Note:}  & \multicolumn{4}{r}{$^{*}$p$<$0.1; $^{**}$p$<$0.05; $^{***}$p$<$0.01} \\ 
\end{tabular} 
\end{table}

\begin{table}[!htbp] \centering 
  \caption{TDprop no interaction terms no scaling.} 
\begin{tabular}{@{\extracolsep{5pt}}lD{.}{.}{-3} D{.}{.}{-3} D{.}{.}{-3} D{.}{.}{-3} } 
\\[-1.8ex]\hline 
\hline \\[-1.8ex] 
 & \multicolumn{4}{c}{\textit{Dependent variable:}} \\ 
\cline{2-5} 
\\[-1.8ex] & \multicolumn{4}{c}{avg\_return} \\ 
\\[-1.8ex] & \multicolumn{1}{c}{BeamRider} & \multicolumn{1}{c}{Breakout} & \multicolumn{1}{c}{SpaceInvaders} & \multicolumn{1}{c}{QBert}\\ 
\hline \\[-1.8ex] 
 lr & 12,624.790 & 1,297.644 & -3,878.601 & -47,169.790^{**} \\ 
  & (17,974.660) & (1,350.642) & (3,599.909) & (18,311.820) \\ 
  & & & & \\ 
 beta & 206.327 & 14.667 & 96.610^{***} & 379.319^{**} \\ 
  & (143.714) & (10.799) & (28.783) & (146.410) \\ 
  & & & & \\ 
 epsilon & 7,921.821 & 79.895 & 1,390.232 & 20,860.190^{***} \\ 
  & (6,601.817) & (496.070) & (1,322.192) & (6,725.653) \\ 
  & & & & \\ 
 Constant & 752.410^{***} & 9.691 & 313.879^{***} & 382.709^{***} \\ 
  & (93.646) & (7.037) & (18.755) & (95.402) \\ 
  & & & & \\ 
\hline \\[-1.8ex] 
Observations & \multicolumn{1}{c}{50} & \multicolumn{1}{c}{50} & \multicolumn{1}{c}{50} & \multicolumn{1}{c}{50} \\ 
R$^{2}$ & \multicolumn{1}{c}{0.095} & \multicolumn{1}{c}{0.066} & \multicolumn{1}{c}{0.213} & \multicolumn{1}{c}{0.285} \\ 
Adjusted R$^{2}$ & \multicolumn{1}{c}{0.036} & \multicolumn{1}{c}{0.005} & \multicolumn{1}{c}{0.161} & \multicolumn{1}{c}{0.238} \\ 
Residual Std. Error (df = 46) & \multicolumn{1}{c}{321.667} & \multicolumn{1}{c}{24.171} & \multicolumn{1}{c}{64.423} & \multicolumn{1}{c}{327.701} \\ 
F Statistic (df = 3; 46) & \multicolumn{1}{c}{1.612} & \multicolumn{1}{c}{1.077} & \multicolumn{1}{c}{4.144$^{**}$} & \multicolumn{1}{c}{6.109$^{***}$} \\ 
\hline 
\hline \\[-1.8ex] 
\textit{Note:}  & \multicolumn{4}{r}{$^{*}$p$<$0.1; $^{**}$p$<$0.05; $^{***}$p$<$0.01} \\ 
\end{tabular} 
\end{table}

\end{document}